\documentclass[]{fairmeta}

\usepackage[utf8]{inputenc}
\usepackage[T1]{fontenc}
\usepackage{lmodern}

\usepackage{microtype}
\usepackage{hyperref}
\usepackage{url}
\usepackage{booktabs}
\usepackage{amsfonts}
\usepackage{nicefrac}
\usepackage{xcolor}

\usepackage{amsmath,amssymb}
\usepackage{graphicx}
\usepackage{enumitem}
\usepackage{placeins}
\usepackage{float}
\usepackage{algorithm}
\usepackage{algpseudocode}
\usepackage{multirow}
\usepackage{xspace}

\hypersetup{
  colorlinks=true,
  linkcolor=blue!50!black,
  citecolor=blue!50!black,
  urlcolor=blue!50!black,
}
 \renewcommand{\hat}{\widehat}

 \renewcommand{\vec}[1]{\ensuremath{\boldsymbol{#1}}}

 \newcommand{\mc}[1]{\ensuremath{\mathcal{#1}}}

 \renewcommand{\eqref}[1]{(\ref{eq:#1})}
 \newcommand{\Eqref}[1]{Equation (\ref{eq:#1})}
 \newcommand{\Figref}[1]{Figure~\ref{fig:#1}}
 
 \newcommand{\tabref}[1]{Table~\ref{tab:#1}}
 \newcommand{\secref}[1]{Section~\ref{sec:#1}}
 \newcommand{\appref}[1]{Appendix~\ref{app:#1}}

\title{Goodbye Drift: Anchored Tree Sampling for Long-Horizon Video-to-Video Generation}
\author[]{Matthew Bendel} 
\author[]{Stephen W. Bailey} 
\author[]{Mithilesh Vaidya} 
\author[]{Sumukh Badam} 
\author[]{Xingzhe He} 

\affiliation[]{Descript, Inc.}

\abstract{
Long-horizon video generation suffers from two intertwined issues. 
First, there is drift, where video quality degrades over time. 
Second, there are continuity issues which manifest as object permanence issues, or improperly rendering transient content (e.g., an object that appears in non-consecutive frames changing color/style).
Recent work has focused on autoregressive distillation techniques that attack both problems simultaneously.
We instead choose to focus on drift directly and introduce \textbf{Anchored Tree Sampling (ATS)}: a training-free inference-time scheduler that replaces left-to-right rollout with sparse-to-dense, anchor-bounded imputation organized as a tree.
A root call produces sparse anchors over the full horizon, recursive refinement generates intermediate anchors, and final leaf spans are synthesized between neighboring anchors.
This reduces the critical path from $K$ sequential rollout steps to $L+1$ tree-hierarchical steps and converts horizon-compounding drift into anchor-bounded drift.
We focus on V2V generation in the \emph{static-camera} regime, where sparse anchors over the horizon are well approximated by the dense conditioning signal, and the base model can produce them without retraining.
We evaluate ATS against two contemporary autoregressive baselines on Wan~$2.1$~$+$~VACE, across five conditioning modalities (inpainting, outpainting, edge, pose, depth).
We show that ATS outperforms both competitors in overall quality, as well as in drift prevention.
We additionally demonstrate stable $\geq 40$-minute generation on LTX-$2.3$ across the same five modalities.
We conclude by proposing a path forward to extend ATS to arbitrarily long T2V generation, as well as the dynamic-camera and multi-shot regimes.
\vspace{-10pt}
}

\metadata[Project Page]{\url{https://descriptinc.github.io/ATS/}}
\metadata[Code]{\url{https://github.com/descriptinc/ATS/}}

\begin{document}
\maketitle

\section{Introduction}
\label{sec:introduction}
Long-horizon video generation faces two related challenges. 
\emph{Drift} manifests as errors which compound over time, degrading video quality.
Second, \emph{continuity} issues manifest as a lack of object permanence, or by improperly rendering spurious content (e.g., inconsistent structure/style of objects that enter/leave the frame at different points in time).
These challenges stem from the fact that video generators have finite context windows, so when the requested output exceeds the window the default approach is to sample autoregressively.
To solve drift and continuity issues, autoregressive distillation techniques \cite{causvid,self_forcing,self_forcing_pp,longlive,diffusion_forcing} are proposed to turn a pretrained bidirectional video model into a causal autoregressive generator.
Despite these advances, autoregressive approaches still suffer from drift, or hard maximum duration constraints \cite{self_forcing_pp} resulting in continuity issues.
Autoregressive sampling also requires sequential rollout, forcing linear growth in wall-clock generation time, even for offline settings where all conditions are known.

Existing long-form video work concentrates overwhelmingly on text-to-video (T2V).
Long-form \emph{video-to-video} (V2V), where a per-frame conditioning signal $\vec{c}_{1:T}$ (mask, edge, pose, depth, etc.) is given over the full horizon and the model must regenerate content faithful to it, is by contrast largely underexplored at length.
Bidirectional V2V conditioning adapters such as VACE~\cite{vace} assume short clips, and the causal-distilled V2V variants we are aware of \cite{vace_ar} inherit the same drift-and-continuity problem as their T2V counterparts once pushed past their training horizon.

\begin{figure}[t]
\centering
\includegraphics[width=\columnwidth]{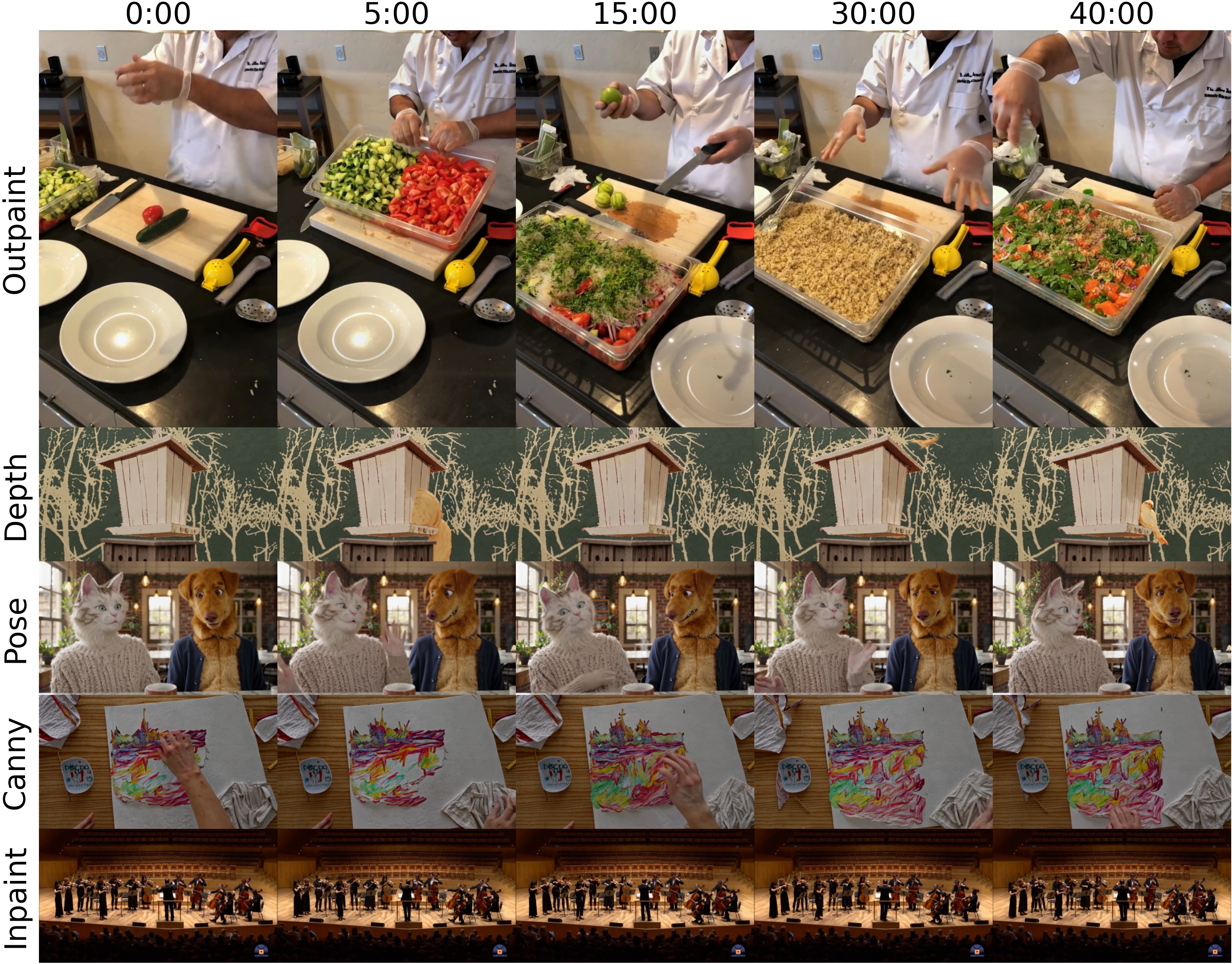}
\caption{Long-horizon V2V with ATS on LTX-$2.3$ in the static-camera regime. ATS eliminates horizon-scaling content drift, holding identity, appearance, and scene state stable across anchor-bracketed spans throughout the full horizon.}
\label{fig:hero}
\end{figure}


We therefore ask a key question: if the full conditioning $\vec{c}_{1:T}$ is known in advance, why insist on causal rollout at all?
Our answer is \textbf{Anchored Tree Sampling (ATS)}, a training-free sampler that attacks drift and continuity issues directly by replacing left-to-right rollout with \emph{sparse-to-dense bounded imputation}.
It assumes only that the base model can generate an interior span while conditioning on information at both its start and end (we call these boundary conditions \emph{anchors}) and organizes long-horizon generation hierarchically.
First, we establish sparse anchors over the full horizon, recursively generate intermediate anchors if needed, then generate dense leaves between neighboring anchors.
This distinguishes our method from causal-distillation approaches that seek to stabilize sequential rollouts but retain a fundamentally horizon-scaling dependency graph~\cite{causvid,lct,causal_forcing,self_forcing,context_forcing,relax_forcing,longlive,diffusion_forcing,long_context_ar_video,memflow,reward_forcing,lpm}, from token-level bidirectional decoding within a single pass~\cite{mask_predict,insertion_transformer,levenshtein_transformer,maskgit,sundae}, and from training-time hierarchical or multi-scale generation that bakes the hierarchy into the model.
ATS does not separately enforce long-range continuity, instead leveraging the base model's strong local-continuity prior within each leaf and inheriting global continuity from the anchor structure.

We focus on the \emph{static-camera} regime in this paper.
The root call requires the base model to emit sparse output at chosen anchor times, but base video generators are trained on continuous footage.
This makes producing sparse outputs a challenge for general video generation.
However, this does not apply for static-camera inputs, as sparse and dense conditioning look near-identical because most of the frame is fixed, so the base model produces clean sparse anchors out-of-the-box.
For dynamic-camera inputs paired with weak motion-control signals, multi-shot or multi-scene content, and pure long-form text-to-video, the same continuity bias overrides the sparse-output request.
One could try to prompt the base model to produce cuts, but without post-training this is not reliable.
Extending ATS to those regimes requires a dedicated sparse-generation specialist alongside the existing dense leaf model, and a generalization of the strictly temporal tree to a directed acyclic graph (DAG).
The closest existing primitive is the keyframe generator of DCARL~\cite{dcarl}, which is the right kind of 
\clearpage
\begin{figure}[t]
    \centering
    \begin{subfigure}[t]{0.9\columnwidth}
        \centering
        \includegraphics[width=\linewidth]{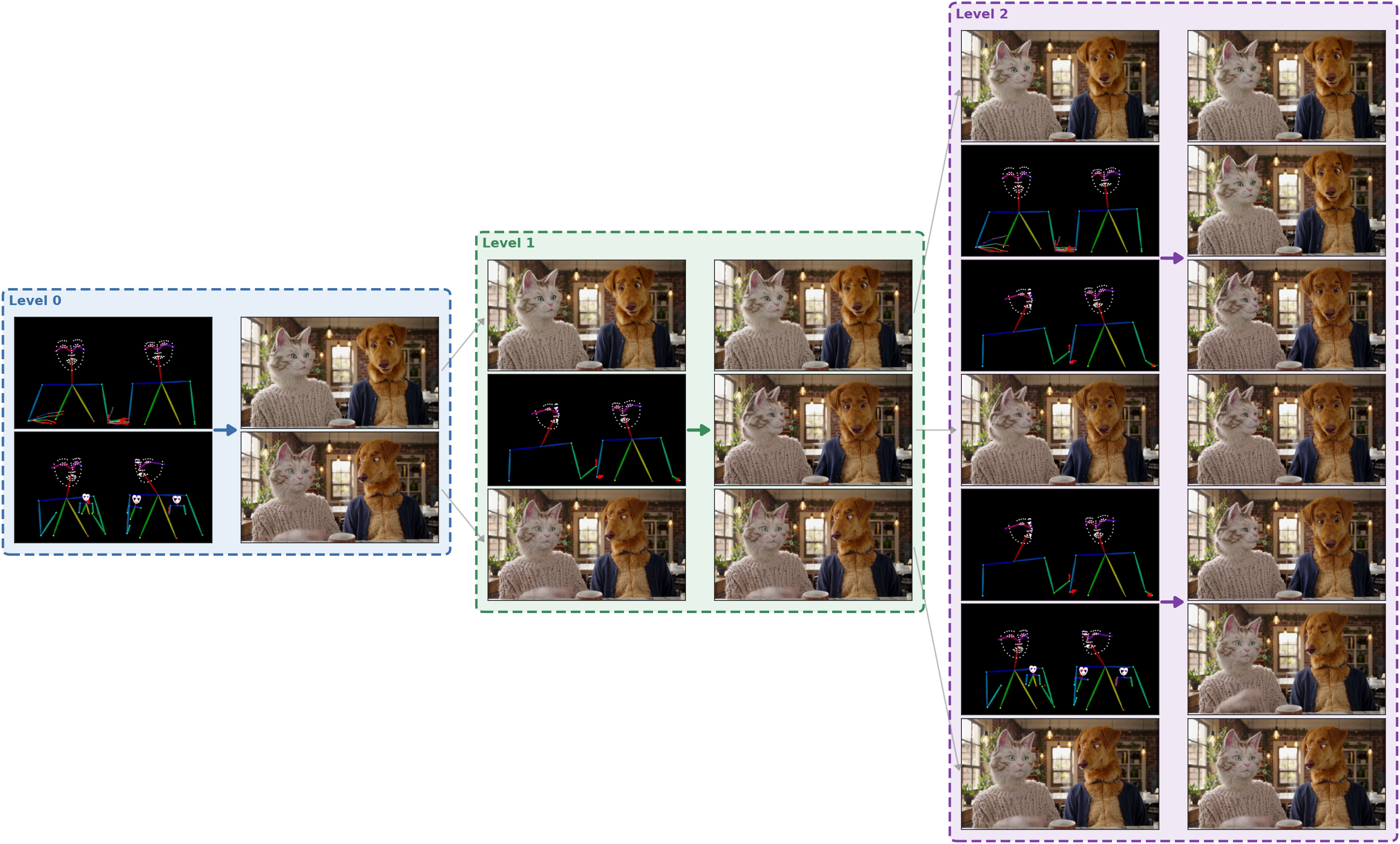}
        \caption{Tree structure. Three level chunks side-by-side, each
        showing the cells visible at that level (cond input on the left,
        gen output on the right of each chunk).}
        \label{fig:tree-structure}
    \end{subfigure}

    \vspace{0.5em}

    \begin{subfigure}[t]{0.9\columnwidth}
        \centering
        \includegraphics[width=\linewidth]{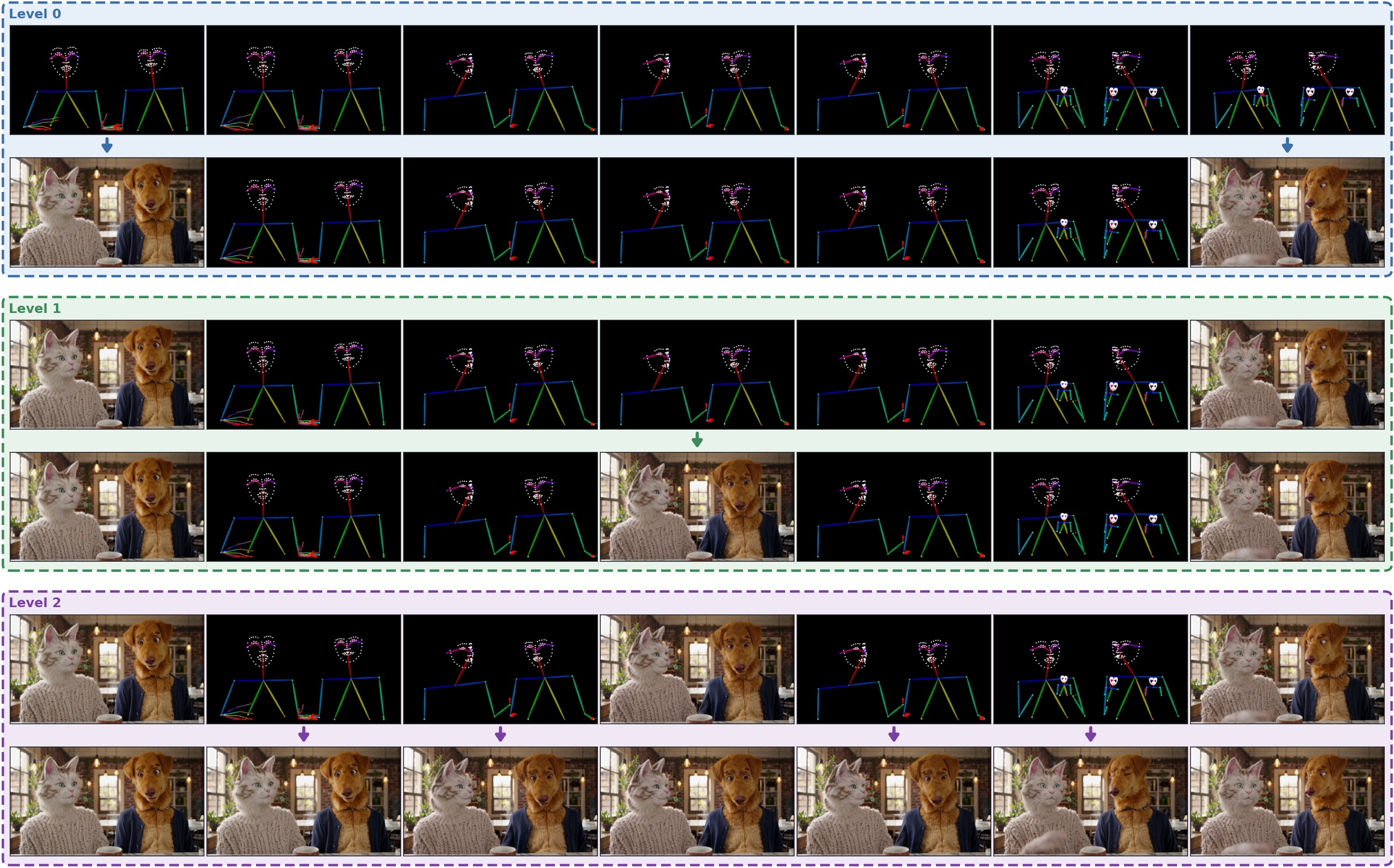}
        \caption{Sparse-to-dense filling. Each chunk shows the full
        timeline twice: input row above (state before this level) and
        output row below (state after).}
        \label{fig:tree-progression}
    \end{subfigure}

    \caption{An overview of our proposed sampling method, ATS. A root call
    produces sparse anchors that span the entire horizon
    \subref{fig:tree-structure}; subsequent guidance and leaf calls
    progressively fill in intermediate anchors and dense neighbour frames
    \subref{fig:tree-progression}.}
    \label{fig:ats-overview}
\end{figure}
\clearpage
sparse model but is composed autoregressively with a dense generator. 
ATS's bidirectional, parallel-sibling tree is the natural replacement for that autoregressive composition.
We discuss the path through these regimes in detail in \secref{path} and leave the corresponding implementation and evaluation to future work.

\Figref{hero} shows ATS at work on LTX-$2.3$~\cite{ltx2} for $\geq 40$-minute generations, covering each modality at both 720p and 1080p.
There, we see that ATS unlocks long-duration generation with no horizon-scaling content drift.
\Figref{ats-overview} illustrates our proposed sampling method. The left panel shows the three-level call hierarchy: a root call generates sparse anchors that span the entire horizon, a guidance call inserts intermediate anchors between them, and leaf calls fill in the remaining dense neighbour frames. The right panel shows the corresponding timeline view, where each level's generator only modifies the cells it is responsible for.
The root level fills the boundary anchors, intermediate levels insert/refine new anchors, and the leaf level produces the dense frames between consecutive anchors.

\section{Background and related work}
\label{sec:background}

We organize this section around the question that motivates our method: when a base generator has only a bounded temporal window, what dependency structures can be used to compose it into long-form output?
We first formalize the standard autoregressive (AR) answer and its failure modes, then describe the two-sided primitive that replaces it, and finally survey existing schedules that compose such a primitive over long horizons and connect them to broader trends in inference-time compute.

\subsection{Autoregressive long-form generation and its failure modes}
\label{sec:bg-ar}

We assume that a V2V generative model produces a target signal $\vec{x}_{1:T}$ aligned with conditioning $\vec{c}_{1:T}$, where $\vec{c}_{1:T}$ represents some per-frame conditioning information.
When $T$ exceeds the model's native context $M_{\max}$, the standard solution is chunked AR sampling. 
The horizon is partitioned into chunks $B_1,\ldots,B_K$ with $|B_k|\leq M_{\max}$, and each chunk is generated according to
\begin{equation}
\label{eq:ar-chunk}
\hat{\vec{x}}_{B_k} \;\sim\; p_\theta\!\left(\vec{x}_{B_k} \,\middle|\, \vec{h}_{k-1},\, \vec{c}_{B_k}\right),
\end{equation}
where $\vec{h}_{k-1}$ is some historical representation of previous chunks.
\Eqref{ar-chunk} is exact only if $\vec{h}_{k-1}$ is a sufficient statistic for the entire history $\vec{x}_{<B_k}$, which is rarely the case for nontrivial signals.
The resulting approximation is responsible for the long-range failure modes that motivate this paper.

The first failure is \emph{drift}, which manifests as quality degradation over time.
The conditioning of chunk $k+1$ is itself a sample from \Eqref{ar-chunk}, so any local distribution shift at chunk $k$ enters the suffix that conditions chunk $k+1$, which in turn enters the suffix conditioning chunk $k+2$, and so on.
Thus, errors compound over time, leading directly to quality degredation.
There is no mechanism in \Eqref{ar-chunk} to suppress drift, because chunk $k$'s output is conditioned only on its past with no future constraint to correct it.
In V2V, the per-frame conditioning track $\vec{c}_{1:T}$ acts as a strong local anchor on the quality axis: each chunk is pinned to a structural reference (mask, edge, depth, pose) and remains locally faithful.
The content axis is not anchored, instead identity and appearance live in $\vec{h}_{k-1}$ and still accumulate error as $\vec{h}_k$ grows.

To stay tractable over long horizons, every recent AR long-form sampler we are aware of maintains a rolling KV cache with periodic refresh, e.g., LongLive~\cite{longlive} uses KV recache plus frame-sink anchors and Self-Forcing / Reward Forcing~\cite{self_forcing,reward_forcing} use chunked rollout with bounded context.
This produces an additional characteristic ``reset'' failure pattern.
Between resets, KV states accumulate error: noise statistics shift, and color, contrast, and texture degrade as content drifts.
At each reset, the model effectively starts a new clip conditioned only on a tiny anchor window (the last $N$ frames, sometimes just one).
However, long-range identity and scene memory are gone, so the rollout produces a discrete content jump which yields different object pose, slightly different palette, sometimes a different object entirely.
This jumpy behavior directly forces AR methods to exhibit continuity issues.

AR methods also suffer from a sequential rollout requirement.
This is a direct consequence of \Eqref{ar-chunk} forcing $\hat{\vec{x}}_{B_1},\dots,\hat{\vec{x}}_{B_{k-1}}$ to be available before $\hat{\vec{x}}_{B_{k}}$ can begin.
Thus, the wall-clock time of a full rollout grows linearly in $K$ regardless of how much hardware is available.

A large body of recent video work attacks both failures while keeping \Eqref{ar-chunk} intact, either by distilling bidirectional teachers into causal students \cite{causvid,lct,causal_forcing,diffusion_forcing}, by training students to cope with their own rollout errors \cite{self_forcing,context_forcing,relax_forcing,longlive}, or by repacking the input-frame context to reduce drift in next-frame prediction \cite{framepack}.
All such methods improve the dependency graph of \Eqref{ar-chunk}, but none of them change it.
Once the task is offline and the full conditioning $\vec{c}_{1:T}$ is known before sampling, there is no reason for the right-hand side of \Eqref{ar-chunk} to omit future information.

\subsection{Two-sided generators and bounded infilling}
\label{sec:bg-twosided}

A natural alternative to \Eqref{ar-chunk} is a primitive that conditions on both past and future context simultaneously.
Given anchors $\vec{A}_a$ and $\vec{A}_b$ at the start and end of an interval $[a,b]$, the two-sided primitive samples
\begin{equation}
\label{eq:bidir}
\hat{\vec{x}}_{a:b} \;\sim\; p_\theta\!\left(\vec{x}_{a:b} \,\middle|\, \vec{A}_a,\, \vec{A}_b,\, \vec{c}_{a:b}\right).
\end{equation}
Several pretrained video generators implement \Eqref{bidir} explicitly.
LTX-$2.3$~\cite{ltx2} natively consumes left and right boundary frames together with a per-frame conditioning track (mask, edge, depth, pose, etc.). 
The unified V2V conditioning adapter VACE~\cite{vace} provides the same primitive on top of other open-weight backbones.
Learned video frame-interpolation methods~\cite{film,rife,ldmvfi} can be read as the limiting case of \Eqref{bidir} in which $\vec{A}_a$ and $\vec{A}_b$ collapse to single frames.
The same primitive recurs in token-level parallel decoders~\cite{mask_predict,insertion_transformer,levenshtein_transformer,sundae,maskgit}.

We treat any generator that supports \Eqref{bidir} as a black-box primitive and ask how to compose many calls to it into a coherent long-form sample.
This separation between the modeling problem and the scheduling problem makes our method training-free and modality-agnostic.

\subsection{Hierarchical scheduling and inference-time compute}
\label{sec:bg-hier}

Composing \Eqref{bidir} over a long horizon requires a schedule for placing anchors and ordering the resulting calls.
Existing systems each commit to a particular hierarchy baked into model design or training, NUWA-XL's diffusion-over-diffusion keyframes, Phenaki's shared-token clip sequencing, Lumiere's space-time U-Net~\cite{nuwaxl,phenaki,lumiere}, or stretch a fixed-window model with overlap tricks as in \cite{freenoise,genlvideo,streamingt2v}.
We treat the hierarchy as an \emph{inference-time} decision.
The depth $L$, branching, and anchor representation are chosen at runtime as a function of $T$ and $M_{\max}$, and no part of the base model is modified.
This places ATS in the broader trend of inference-time scheduling as a first-class axis~\cite{tot,snell_test_time}, but applied to long-horizon temporal generation.
The additional compute is spent on coarse-to-fine anchor planning that constrains and parallelizes the dense phase.
For V2V, two regimes follow naturally.
In \emph{source-conditioned editing} a dense source signal supplies local fidelity while the tree supplies global consistency (the natural setting for inpainting, outpainting, edge, depth, pose).
In \emph{reference-conditioned generation} long-range consistency must propagate through sparse anchor structure rather than through a full source trace~\cite{nuwaxl,phenaki,longlive}.

Recently, \cite{dcarl} proposed DCARL, a system which trains a dedicated sparse-keyframe generator alongside a dense generator and composes them in a divide-and-conquer pipeline.
The keyframe generator emits a full keyframe track over the long horizon, after which the dense generator fills in the inter-keyframe spans.
DCARL's two-tier structure matches the sparse-then-dense decomposition we adopt, but composes its two stages autoregressively, while ATS composes them bidirectionally and in parallel.
We discuss DCARL further in \secref{ats-t2v}.

\section{Anchored Tree Sampling}
\label{sec:tree}

We now describe how the two-sided primitive of \Eqref{bidir} can be composed into a long-form sampler.
The construction reorganizes long-horizon generation as a tree of bounded infilling problems.
The resulting dependency graph is top-down rather than left-to-right.

\subsection{A hierarchy of anchors}
\label{sec:tree-setup}

We retain the notation of \secref{bg-ar}, where a target $\vec{x}_{1:T}$ is generated with time-aligned conditioning $\vec{c}_{1:T}$, and the base model is treated as a black-box implementation of \Eqref{bidir} subject to a maximum interval length $M_{\max}$ and a minimum coherent length $M_{\min}$.
An \emph{anchor} $\vec{A}_t$ is any boundary representation the model can consume in place of $\vec{x}_t$ (a short clip or a keyframe).
What matters is that anchors sit at known temporal positions, are consumable as start or end conditions in \Eqref{bidir}, and carry enough local identity and dynamics information to keep the resulting infilling problem bounded.

The sampler organizes generation around a nested family of anchor sets
\begin{equation}
\label{eq:hierarchy}
\mc{A}^{(0)} \;\subseteq\; \mc{A}^{(1)} \;\subseteq\; \cdots \;\subseteq\; \mc{A}^{(L)},
\qquad
\mc{A}^{(\ell)} \;=\; \bigl\{\vec{A}_{t^{(\ell)}_0}, \vec{A}_{t^{(\ell)}_1}, \ldots, \vec{A}_{t^{(\ell)}_{N_\ell}}\bigr\},
\end{equation}
with $t^{(\ell)}_0 = 1$ and $t^{(\ell)}_{N_\ell} = T$ at every level.
The coarsest set $\mc{A}^{(0)}$ covers the full horizon, and each finer level inserts new anchors strictly between adjacent anchors of the previous level, so coarser anchors persist at all finer levels.
The leaves of the tree are the dense intervals enclosed by adjacent anchors at the finest level $\mc{A}^{(L)}$, each of length in $[M_{\min}, M_{\max}]$.
Depth $L$ is determined entirely by the relationship between $T$, the effective context span of the chosen sparse representation, and the dense leaf budget.
The root call produces $\mc{A}^{(0)}$ to seed the tree, and $L=0$ suffices when the resulting consecutive level-$0$ anchors are spaced within $[M_{\min}, M_{\max}]$ so leaf calls can fill them in directly.
Deeper $L$ is added recursively whenever any level-$\ell$ interval still exceeds $M_{\max}$, until every node call fits within the base model's context budget.
No modality-specific heuristic is involved.

\subsection{Top-down factorization and parallel inference}
\label{sec:tree-inference}
Generation proceeds in three structurally distinct stages.
The root stage produces the coarsest anchor set $\mc{A}^{(0)}$ in a single \emph{conditioning-only} call to $p_\theta$, emitting the level-$0$ anchors at the sparse times $t^{(0)}_0, \ldots, t^{(0)}_{N_0}$ from a sparse view of $\vec{c}_{1:T}$ alone (e.g.\ sparsely sampled source frames for source-conditioned video, or position-matched references for reference-conditioned video).
Concretely, given a root span $[t_\text{start}, t_\text{end})$ and VAE temporal stride $s_f$ ($s_f{=}8$ for LTX-$2.3$), we assemble the root input by concatenating along the time axis a lead frame at $t_\text{start}$, an $s_f$-frame slice centered on each interior anchor time, and a final $s_f$-frame slice at the right boundary; the task-specific applier (mask / Canny / depth / pose) is mapped pointwise across this clip so the same source frame produces identical conditioning wherever it appears.
What is extracted from each decoded window depends on whether motion continuity is supplied externally: source-conditioned settings (inpainting, outpainting, edge, pose, depth) already pin down cross-boundary motion via the per-frame conditioning track, so a single keyframe per window suffices, while reference-conditioned settings retain the full $s_f$-frame window as a multi-frame anchor clip so descendants inherit local motion context.
\Eqref{bidir} accommodates either; the five modalities in \secref{experiments} are all source-conditioned and use the keyframe variant.
The refinement stage then generates each new anchor at level $\ell$ via an instance of \Eqref{bidir} whose start and end conditions are the enclosing pair of level-$(\ell{-}1)$ anchors.
Writing $I^{(\ell)}_i = (t^{(\ell)}_i, t^{(\ell)}_{i+1})$, $\Delta^{(\ell)}_i$ for the level-$\ell$ anchors strictly inside the $i$-th level-$(\ell{-}1)$ interval, and $\hat{\vec{x}}^{(L)}_j$ for the leaf-level dense content inside the $j$-th level-$(L{-}1)$ interval, the sampler realizes the joint
\begin{align}
\label{eq:tree-factor}
p_\theta(\hat{\vec{x}}_{1:T}) \;=\;
&\underbrace{p_\theta\!\left(\mc{A}^{(0)} \,\middle|\, \vec{c}_{1:T}\right)}_{\text{root}}
\;\cdot\;
\prod_{\ell=1}^{L-1} \prod_{i} p_\theta\!\left(\Delta^{(\ell)}_{i} \,\middle|\, \vec{A}_{t^{(\ell-1)}_i}, \vec{A}_{t^{(\ell-1)}_{i+1}},\, \vec{c}_{I^{(\ell-1)}_i}\right)\nonumber\\
&\;\cdot\;
\prod_{j} p_\theta\!\left(\hat{\vec{x}}^{(L)}_j \,\middle|\, \vec{A}_{t^{(L-1)}_j}, \vec{A}_{t^{(L-1)}_{j+1}},\, \vec{c}_{I^{(L-1)}_j}\right),
\end{align}
corresponding to the root call, internal anchor refinement, and dense leaf generation.
Every factor after the root is exactly an instance of \Eqref{bidir}, and the root is the same generator in its no-anchor mode, so any pretrained two-sided generator serves as the base model without modification.
Sibling factors at a common level are conditionally independent given their parents; the only sequential dependence is root\,$\to$\,refinement\,$\to$\,leaves.

Algorithm~\ref{alg:tree} defines the ATS procedure.
\begin{algorithm}[t]
\caption{Anchored Tree Sampling (ATS).}
\label{alg:tree}
\begin{algorithmic}[1]
\Require horizon $T$, conditioning $\vec{c}_{1:T}$, base model $p_\theta$ implementing \Eqref{bidir}, limits $(M_{\min},M_{\max})$
\Ensure sample $\hat{\vec{x}}_{1:T}$
\State Plan tree depth $L$ and anchor times $\{t^{(\ell)}_i\}_{\ell,i}$ so every leaf-level interval has length in $[M_{\min},M_{\max}]$
\State $\mc{A}^{(0)} \;\sim\; p_\theta\!\left(\,\cdot\,\middle|\, \vec{c}_{1:T}\right)$ at sparse times $t^{(0)}_0, \ldots, t^{(0)}_{N_0}$ \Comment{root: conditioning only, no anchors}
\For{$\ell = 1, \ldots, L$} \Comment{coarse-to-fine}
    \ForAll{adjacent anchor pairs $(\vec{A}_a, \vec{A}_b) \in \mc{A}^{(\ell-1)}$ \textbf{in parallel}}
        \State $\hat{\vec{x}}^{(\ell)}_{a:b} \;\sim\; p_\theta\!\left(\,\cdot\,\middle|\, \vec{A}_a,\, \vec{A}_b,\, \vec{c}_{a:b}\right)$ \Comment{bidirectional sample via \Eqref{bidir}}
    \EndFor
    \If{$\ell < L$}
        \State $\mc{A}^{(\ell)} \gets \mc{A}^{(\ell-1)} \cup \bigl\{\text{anchors extracted from each } \hat{\vec{x}}^{(\ell)}_{a:b}\bigr\}$
    \EndIf
\EndFor
\State \Return $\hat{\vec{x}}_{1:T}$ assembled from $\bigl\{\hat{\vec{x}}^{(L)}_{a:b}\bigr\}$ in temporal order \Comment{leaves are the level-$L$ outputs}
\end{algorithmic}
\end{algorithm}
Three properties follow.
No leaf extrapolates open-endedly, since every dense interval has both a left and right anchor; sibling generations are runtime-independent and execute concurrently; and the sampler never modifies the base model, so the construction is training-free.
ATS thereby separates two roles entangled in AR rollout: coarse levels of \Eqref{tree-factor} carry global structure (aiding continuity) while leaves solve local bounded-infilling (eliminating drift).

\subsection{Compute, latency, and drift}
\label{sec:tree-analysis}

The systems consequence of \Eqref{tree-factor} concerns latency rather than total compute.
The tree organizes generation into $L+1$ steps, featuring a root call followed by $L$ subsequent levels (the last of which produces dense leaves.
Each level of the tree consists of conditionally independent sibling calls, so under sufficient parallel hardware the critical-path latency reduces to
\begin{equation}
\label{eq:latency}
\mc{T}_{\text{tree}} \;\approx\; (L+1)\,\tau_{\text{call}},
\qquad
\mc{T}_{\text{AR}} \;\approx\; K\,\tau_{\text{call}},
\end{equation}
where $\tau_{\text{call}}$ is the cost of a single base-model call, $K = \lceil T/M_{\max}\rceil$ is the AR chunk count, and $L<<K$ as $T$ grows.
Once a sparse representation is fixed, $L$ scales logarithmically in $T$ while $K$ scales linearly, so the tree converts horizon-scaling latency into depth-scaling latency.

By construction, all non-root nodes produced by ATS are performing bounded infilling.
Consequently, whether or not there is drift is fully dictated by how well the model respects the anchors used to guide it.
Similarly, the hierarchical factorization mitigates (but does not eliminate) continuity issues.
ATS produces perfectly consistent content \emph{when everything is present in the root node}.
However, this means that a poor level-$\ell$ anchor corrupts all of its descendants.
Note that a failed anchor is different than a single bad leaf, which can be regenerated in isolation due to the conditional independence of all sibling nodes.

\section{Experimental Results}
\label{sec:experiments}

\begin{figure}[t]
\centering
\includegraphics[width=0.7\columnwidth]{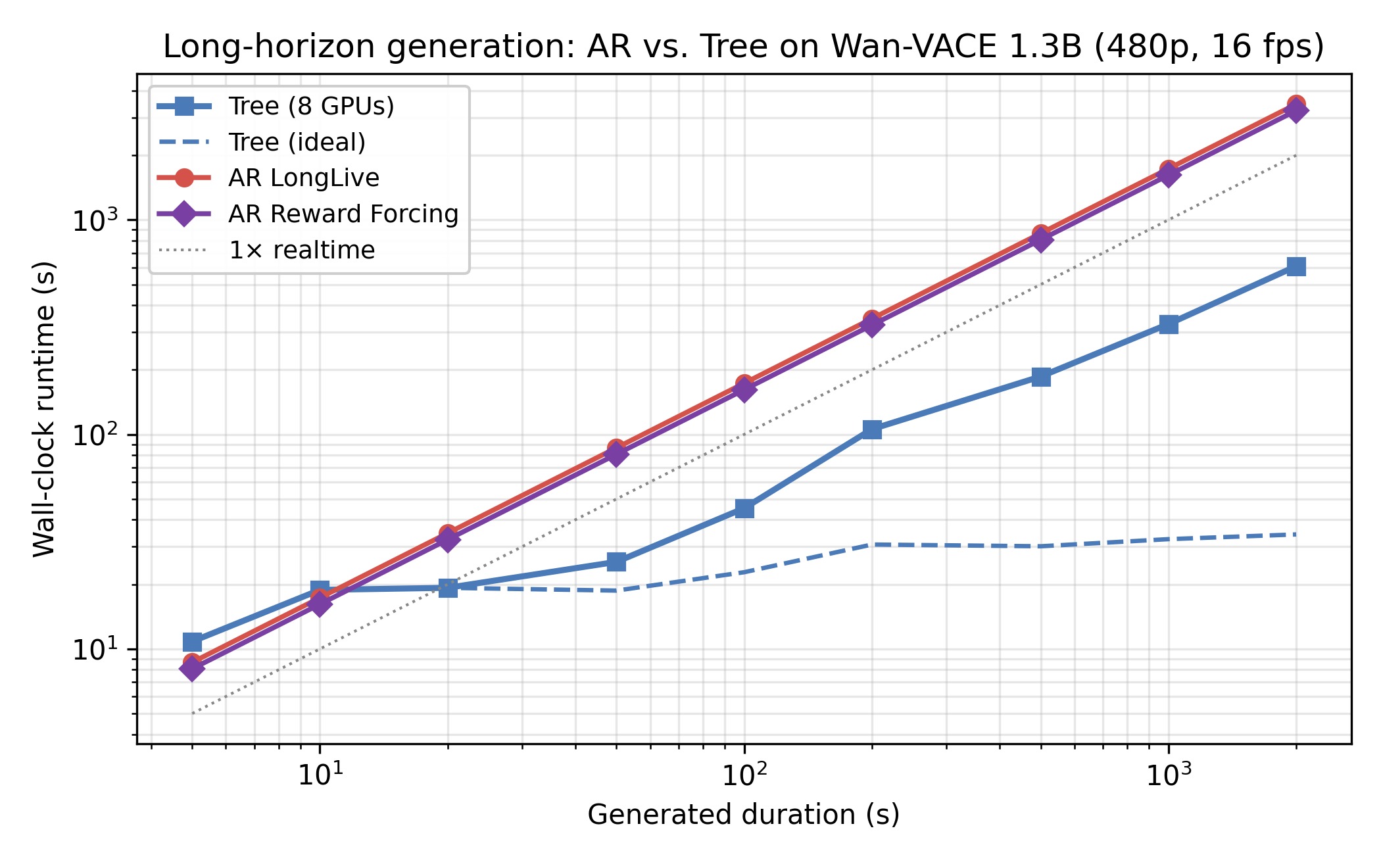}
\caption{Runtime vs.\ generated duration on Wan~$2.1$~$+$~VACE. ATS' parallelism allows it to achieve faster-than-realtime performance.}
\label{fig:runtime}
\end{figure}

\begin{figure}[t]
\centering
\includegraphics[width=\columnwidth]{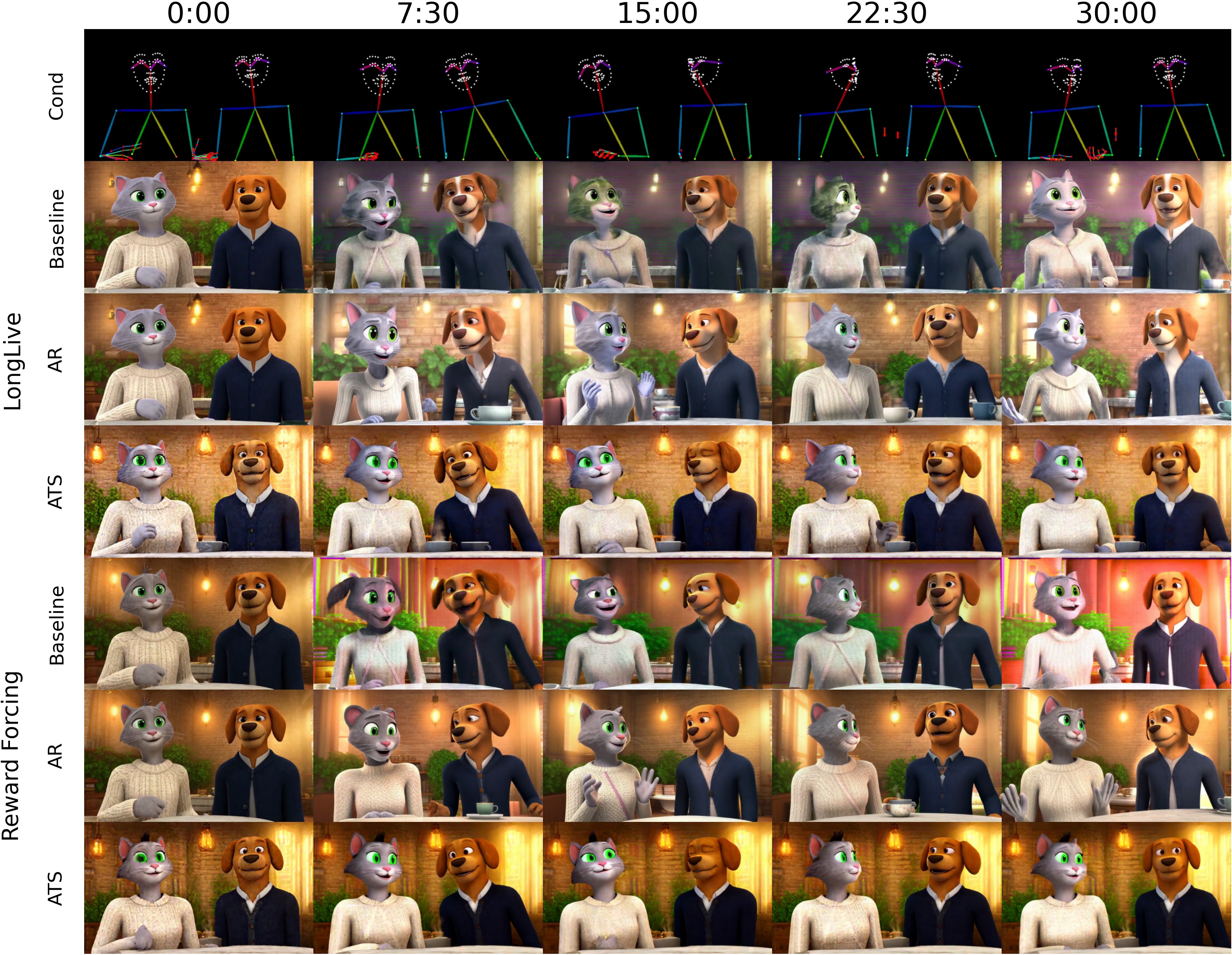}
\caption{Long-form pose-conditioned generation on Wan~$2.1$~$+$~VACE: AR vs.\ ATS on the same checkpoints.}
\label{fig:pose}
\end{figure}

\begin{figure}[t]
\centering
\includegraphics[width=\columnwidth]{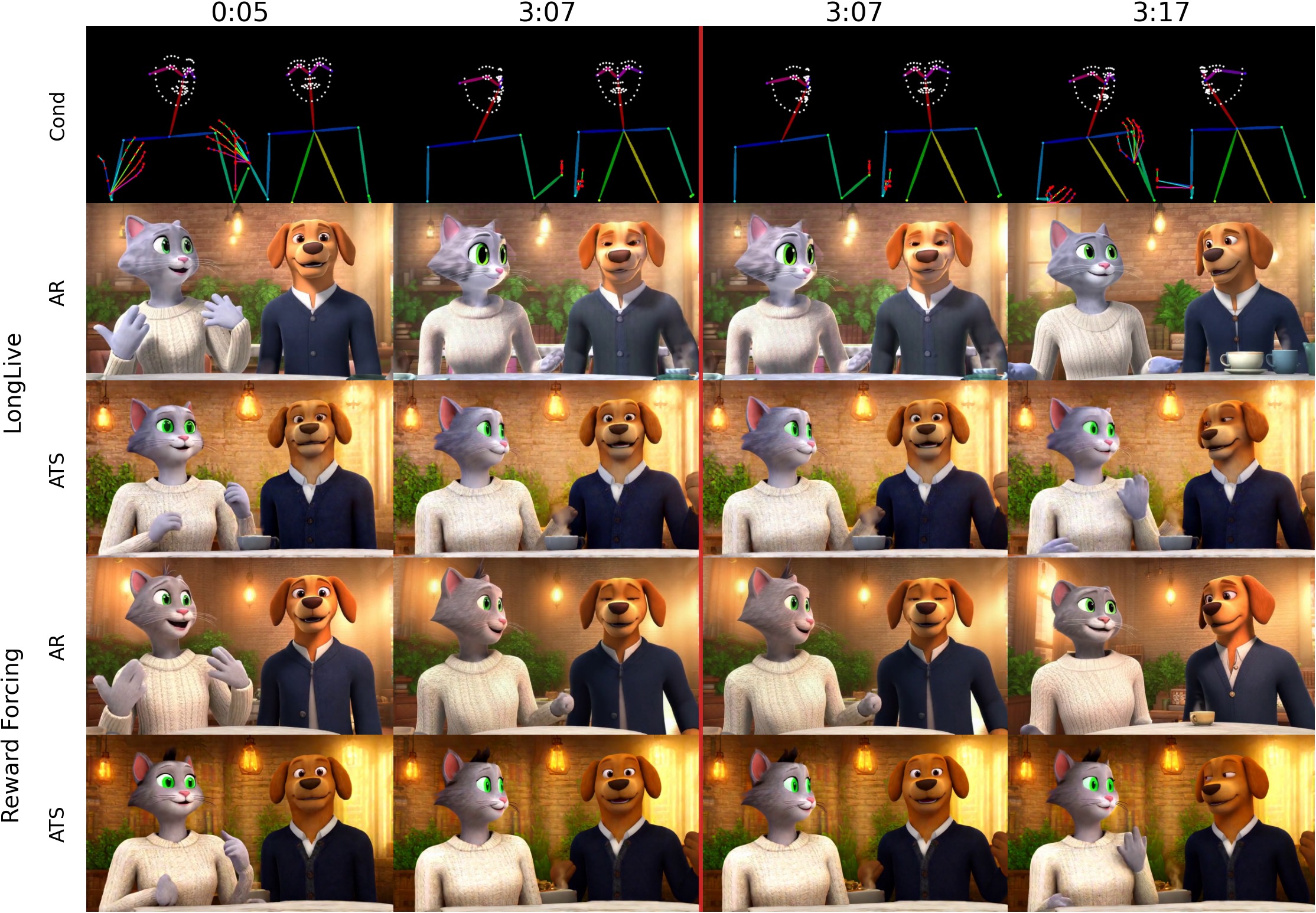}
\caption{Qualitative drift and cache-reset visualization for pose-guided Wan~$2.1$~$+$~VACE generation. The left pair shows intra-chunk drift accumulated within a single cache window, while the right pair illustrates the cross-reset jump.}
\label{fig:cache_pose}
\end{figure}

We restrict the empirical demonstration to the static-camera regime defined in \secref{introduction}.

\paragraph{Setup.}
We evaluate ATS against two contemporary autoregressive long-form V2V baselines on Wan~2.1 $+$ VACE~\cite{wan21,vace,vace_ar}, the standard open-weight bidirectional V2V backbone.
The baselines we consider are LongLive~\cite{longlive} and Reward Forcing~\cite{reward_forcing}.
Each distill Wan~2.1~$+$~VACE into a causal sampler.
For ATS, we use the same distilled checkpoint as the leaf generator and the unmodified Wan~2.1~$+$~VACE bidirectional base for the root call.
Each distilled checkpoint therefore appears twice.
Once rolled out autoregressively, and once invoked inside the tree.
This isolates the scheduler from the model and produces four configurations.

We evaluate five source videos, one per VACE conditioning modality considered (inpainting, outpainting, edge, pose, depth), each trimmed to a $30$-minute horizon with $M_{\min} = 33$ and $M_{\max}=81$ frames for ATS, although ATS supports substantially longer horizons (see the $\geq 40$-minute LTX-$2.3$ reels in \Figref{hero} and \appref{ltx-full}).
The source clips are listed in \appref{eval-set}.
All results are generated with a fixed seed and we used DepthAnything V3 \cite{depthanything3} for depth maps, Canny \cite{canny} for edge maps, and DWPose \cite{dwpose} for pose video.
We evaluate using two no-reference quality predictors aggregated across the five videos: Aesthetic Quality (AQ) and Imaging Quality (IQ), both VBench-$1.0$ indicators~\cite{vbench}.
Letting $M(\cdot)$ denote either predictor applied to a video window, each predictor is reported in three regimes: (i) the \emph{global} score, mean over $60$ uniformly-sampled keyframes, (ii) the \emph{chunk drift} $|\Delta M|_c$, mean over the ten cache-reset windows of $|M(s_i) - M(e_i)|$ where $s_i, e_i$ are samples taken $5$\,s after and before consecutive resets, and (iii) the \emph{reset jump} $|\Delta M|_r$, the same protocol applied across reset boundaries, $|M(e_i) - M(s_{i+1})|$.

\paragraph{Global quality.}
Table~\ref{tab:vbench} reports mean AQ and IQ for all methods under test. 
ATS sustains higher per-frame quality than AR on both checkpoints, quantified as $+4.0$ AQ / $+6.6$ IQ for LongLive, and $+5.7$ AQ / $+6.2$ IQ for Reward Forcing. 
Over the full $30$-minute horizon, the AR rollout has shed $\approx 6$ IQ points relative to ATS scheduled on the same underlying generator.

\paragraph{Chunk drift: gradual within-cycle drift.}
Because each AR pipeline performs a hard KV-cache reset every $\approx 192$\,s, an AR rollout over our $30$-minute horizon is structurally a chain of $10$ independently-conditioned chunks separated by $9$ reset boundaries.
We use this to isolate KV-cache drift from natural source motion.
For each chunk $i$, we sample one frame $5$\,s after the post-reset settle ($s_i$) and one frame $5$\,s before the next reset ($e_i$), and report the per-chunk drift $|\Delta M|_c = |M(s_i) - M(e_i)|$ averaged across the $10$ chunks.
The results are reported in \tabref{vbench}.
The two AR checkpoints behave very differently here.
LongLive drifts heavily within each chunk, while Reward Forcing is more stable. 
However, ATS flattens both, equalizing the chunk quality for both checkpoints.

\paragraph{Reset jump: discrete content discontinuities.}
In-chunk stability is necessary but not sufficient for long-form continuity. 
The cache reset itself is a hard event in which the AR generator discards its accumulated state. 
We measure this discontinuity using the same scalar drift protocol applied across reset boundaries. The results, also reported in \tabref{vbench}, show that jump is large for \emph{both} AR checkpoints, even Reward Forcing which holds quality steady within each chunk.
AR is therefore bounded by at least one of intra-chunk drift or reset-induced discontinuity. 
Reward Forcing only escapes the first, while ATS solves both.

\begin{figure}[t]
\centering
\includegraphics[width=\columnwidth]{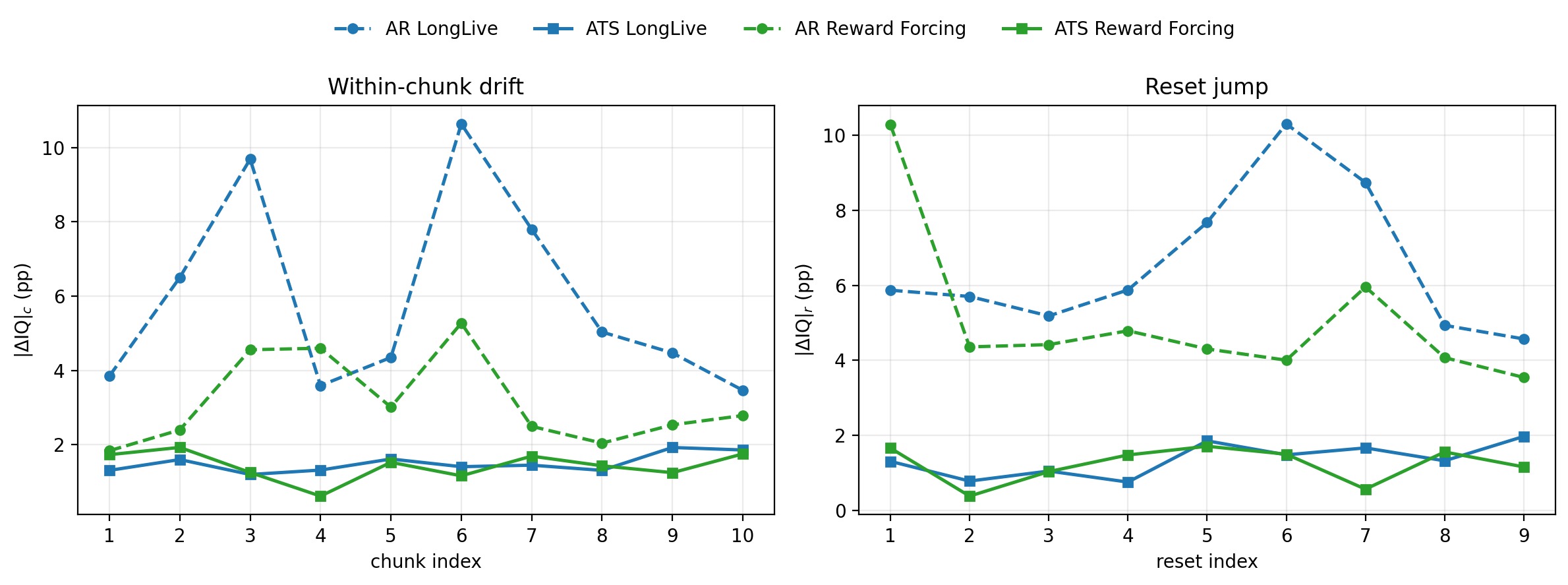}
\caption{Left: $|\Delta\mathrm{IQ}|_c$ for the four methods under test, reported for each chunk. Right: $|\Delta\mathrm{IQ}|_r$ for the four methods under test, reported for each reset boundary. ATS stabilizes both forms of drift.}
\label{fig:iq_drift}
\end{figure}

\Figref{iq_drift} reports the IQ drift metrics for both chunk drift and the reset jump for individual chunks and reset boundaries, respectively. 
There, we see that ATS remains flat over the entire duration of the video, eliminating both forms of drift.

\paragraph{Qualitative reels.}
Figure~\ref{fig:pose} shows pose-guided reels for all four method/sampler combinations sampled uniformly across the full $30$-minute clip.
We also visualize a baseline version of each AR method under a no-cache-reset strategy, in which the hard KV-cache reset is replaced by a position shift: once the RoPE position counter approaches the precomputed table limit, every cached K vector is re-rotated by $R(-N)$ on its temporal channels and the position counter is decremented by $N$, so the cache retains its contents and all relative positions are preserved. This removes the periodic discontinuity but, as shown in \Figref{pose}, quality continues to degrade over time.
Figure~\ref{fig:cache_pose} isolates the two underlying failure modes in a single cache-reset window: the left pair shows intra-chunk drift, and the right pair shows the cross-reset jump. 
LongLive suffers from both issues.
Identity and shading drift visibly across the left pair, and the reset then snaps the scene to a distinct post-reset state on the right. 
Reward Forcing has lower intra-chunk drift across the left pair but still exhibits the reset jump on the right. 
ATS exhibits neither pattern under either checkpoint. 
The remaining four videos appear in Appendix~\ref{app:additional_figs}.

\begin{table}[t]
\centering
\caption{Long-form quality and drift on Wan~2.1+VACE, averaged across the five 30-min source videos. Each distilled checkpoint is run twice: rolled out autoregressively (AR), and scheduled inside the tree (ATS). Global scores are the mean of $60$ uniformly-sampled keyframes. Chunk drift $|\Delta M|_c$ is the mean over the ten cache-reset windows of $|M(s_i) - M(e_i)|$ with $s_i,e_i$ sampled $5$\,s after / before each reset. Reset jump $|\Delta M|_r$ is $|M(e_i) - M(s_{i+1})|$ averaged across the $9$ reset boundaries.}
\label{tab:vbench}
\resizebox{0.75\columnwidth}{!}{%
\begin{tabular}{ll cc cc cc}
\toprule
& & \multicolumn{2}{c}{Global $\uparrow$}
  & \multicolumn{2}{c}{Chunk drift $\downarrow$}
  & \multicolumn{2}{c}{Reset jump $\downarrow$} \\
\cmidrule(lr){3-4} \cmidrule(lr){5-6} \cmidrule(lr){7-8}
Checkpoint & Sampler & AQ & IQ
  & $|\Delta\mathrm{AQ}|_c$ & $|\Delta\mathrm{IQ}|_c$
  & $|\Delta\mathrm{AQ}|_r$ & $|\Delta\mathrm{IQ}|_r$ \\
\midrule
\multirow{2}{*}{LongLive~\cite{longlive}}
 & AR  & 55.69 & 61.82 & 3.73 & 5.94 & 4.32 & 6.54 \\
 & ATS & \textbf{59.69} & \textbf{68.46} & \textbf{2.51} & \textbf{1.50} & \textbf{1.77} & \textbf{1.35} \\
\midrule
\multirow{2}{*}{Reward Forcing~\cite{reward_forcing}}
 & AR  & 52.78 & 63.48 & 3.37 & 3.15 & 3.85 & 5.08 \\
 & ATS & \textbf{58.51} & \textbf{69.64} & \textbf{2.44} & \textbf{1.43} & \textbf{2.08} & \textbf{1.23} \\
\bottomrule
\end{tabular}%
}
\end{table}

\paragraph{Runtime.}
\Figref{runtime} reports runtime vs. generate duration with Wan 2.1 + VACE on the two AR methods under test, ATS with 8 GPUs, and the theoretical-best (one-GPU-per-sibling) parallel configuration that ATS targets.
The two AR baselines scale linearly in $T$ since chunk $i{+}1$ cannot start before chunk $i$ finishes, while ATS scales logarithmically.
A realistic shared-GPU budget shrinks the absolute gap but still maintains a notable speed advantage over the AR baselines.
With just 8 GPUs, ATS is 3.3x faster than realtime and 5.3x faster than the best AR method when generating a 2000s long video.
 
\subsection{Limitations of ATS}
\label{sec:lim-observed}

In all five cases, ATS produces an extremely long video with no horizon-scaling content drift.
However, we observe a number of limitations with ATS, which limit its ability to be used broadly.
Here, we discuss these limitations, then in \secref{path} we propose a concrete path forward to general-purpose long-content generation.
We leave further development of the framework, including exploration of the proposed solutions, to future work.
 
\begin{figure}
    \centering
    \includegraphics[width=0.6\columnwidth]{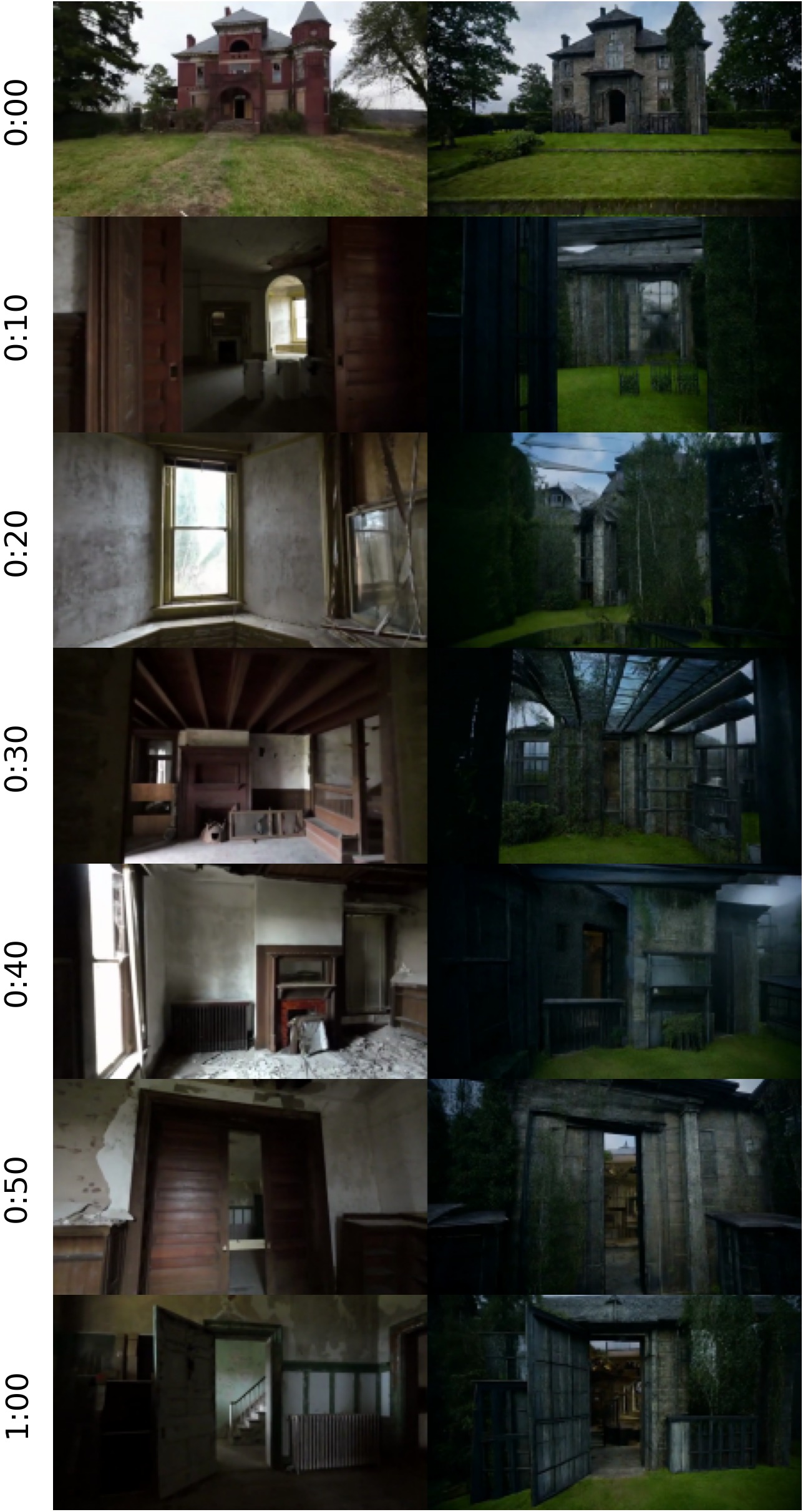}
    \caption{Applying ATS to a 1 minute long drone fly-through video. On the left we show frames from the original source video, on the right we show the corresponding frame produced by ATS with depth conditioning in the root node of the tree. There, we see that ATS fails. The root guidance overfits to the style defined by the first frame, failing to generate sparse content which respects the depth conditioning.}
    \label{fig:depth_fail}
\end{figure}

\paragraph{1.\ A bad anchor poisons its entire subtree.}
Due to the hierarchical propagation of anchors with ATS, a single bad anchor results in degraded quality for all nodes which descend from it.
Consequently, a poor root node result can seriously degrade an entire subtree.
For static camera video, we have anecdotally observed this to not be a significant issue.
However, when trying to apply ATS in its current form to dynamic camera video (or more complicated multi-shot video), generated guidance does not sufficiently adhere to the conditioning in weaker regimes (depth/Canny/pose).
This leads to malformed anchors that overfit to an initial or final condition.
\Figref{depth_fail} illustrates this with a dynamic-camera video generated with Wan2.1 + VACE.
There, we see that root anchors are overfit to the initial conditions of the generated video, and fail to produce coherent sparse content.
Note that a failed leaf generation does not pose the same risk.
Due to the conditional independence of leaves, they can be regenerated individually should they fail.

\paragraph{2.\ Weak motion guidance demands video anchors, not keyframes.}
Unsurprisingly, when motion guidance is weak, keyframes are no longer sufficient to produce smooth motion continuation between chunks.
In these instances, the anchors should consist of small chunks of video.
The effects of this can be seen most prominently in the inpainting video featured in \Figref{inpaint}.
There, some chunks exhibit non-smooth motion continuation (e.g., unnatural movement given the previous chunk).
 
\paragraph{3.\ ATS does not necessarily solve consistency.}
By construction, ATS delivers complete temporal \emph{coherence}, eliminating the \textbf{horizon-scaling content drift} that AR rollouts exhibit under V2V conditioning.
However, due to the independent rendering of all sibling nodes at each level of the tree, ATS does not guarantee consistency for entities outside the support of any bracketing anchor pair.
Specifically, an entity that is out-of-frame in both bracketing anchors of a leaf but appears mid-leaf inherits no identity track from the tree, and the model is free to render it however the local conditioning permits.
This can be seen in \emph{all 5} generated videos.
Each instance of an off-camera object or feature entering frame can result in it having a slightly different appearance (e.g., hands).
This also leads to transient content, where spurious objects can enter frame, or vanish, within one subtree.

\paragraph{4.\ ATS does not provide a sparsity mechanism for non-V2V generation.}
Critically, ATS in its current form is not capable of generalizing to the more general class of non-V2V generation problems (e.g., T2V).
The root and guidance nodes all leverage sparsity from a video conditioning signal.
Without this, one cannot reliably generate a sparse representation of the desired video in order to leverage ATS-style inference.

\subsection{Mitigating the limitations of ATS}
Several smaller-scope mitigations follow directly from the structure of \Eqref{tree-factor} and are useful even before the structural changes we propose in \secref{path}.

\begin{enumerate}
    \item \emph{Adaptive anchor placement} replaces uniform level-$0$ spacing with placement biased toward regions of $\vec{c}_{1:T}$ with high local information content, shifting anchor density toward the regions of the horizon where transient events are most likely.
    \item \emph{Conditioning-residual gating} re-extracts the conditioning track from each leaf output and compares to the input track (e.g., re-extract Canny from the rendered leaf and compare to the input edge map). Large residuals flag intervals where the model has departed from $\vec{c}$, allowing one to rerender errant leaves or even entire corrupted subtrees.
    \item \emph{Cross-leaf consistency} adds a lightweight post-hoc pass that compares adjacent leaf outputs at their shared anchor and resolves residual discontinuities via a short bidirectional inpainting pass at the seams.
\end{enumerate}

None of these change \Eqref{tree-factor}, and none of them substitute for the structural extensions outlined in \secref{path}.
However, these improvements allow ATS to be deployed as is in regimes where static cameras are common (e.g., talking-head videos).

\section{Solving the failures: a path beyond static cameras}
\label{sec:path}

The limitations of \secref{lim-observed} map to three structural changes, none of which alter ATS's parallel-sibling topology.
Limitations~1 and~4 both stem from the same fact: base video generators carry a continuity prior that resists sparse-anchor outputs, so the root call fails in dynamic and T2V regimes.
A dedicated sparse-generation specialist alongside the dense generator (\secref{ats-t2v}) resolves both.
Limitation~3 (consistency) requires more: identity, environment, and object structure must be enforced across non-adjacent leaves, which is the natural fit for a directed acyclic graph (DAG) indexed by semantic rather than temporal subgraphs (\secref{ats-semantic}).
The connecting thread is reference-based conditioning (\secref{ats-references}): anchors are already references in everything but name.
Limitation~2 (weak motion guidance) is the only one not requiring a structural change---ATS's multi-frame anchor variant from \secref{tree-inference} handles it directly.

\subsection{Applying ATS to T2V generation}
\label{sec:ats-t2v}

Pretrained video models are trained on continuous video footage, so sparse outputs are out of distribution and the model cannot reliably produce them.
This \emph{continuity prior} is what makes generalizing ATS to T2V and non-static cameras non-trivial: the root call asks the model to emit content at sparsely chosen times with no continuous frames in between, but the prior insists on a continuous interpolant.
The fix is a new model that produces coherent temporal content at any chosen level of sparsity.

The natural sparsity ladder has three rungs.
\emph{Intra-shot} (dense): consecutive frames inside a single continuous take, where the continuity prior is exactly right.
\emph{Intra-scene} (semi-sparse): representative frames across multiple shots in a single setting and character configuration, where the prior is right at small scale and wrong at the cuts.
\emph{Intra-full-video} (very sparse): representative frames across the full horizon, spanning environment, identity, lighting, and shot changes.
Today's dense generators occupy the bottom; nothing currently occupies the top.

The closest existing primitive is the keyframe generator of DCARL~\cite{dcarl}, trained explicitly to emit a sparse keyframe sequence over a long horizon.
DCARL composes its keyframe and dense generators \emph{autoregressively}: the keyframe generator emits a full keyframe track, and the dense generator fills in left-to-right between consecutive keyframes.
With ATS, the keyframe generator becomes the root call, and every inter-keyframe interval becomes a leaf or small subtree generated bidirectionally and in parallel.
The keyframe generator resolves limitations~1 and~4; ATS contributes the bidirectional, parallel composition DCARL's autoregressive pipeline lacks.
DCARL itself currently assumes intra-shot sparsity (e.g., car-driving POV videos), so the full ladder is not yet realized in any released model.

Even with a working keyframe generator, keyframes alone are unlikely to produce smooth motion continuity between neighboring chunks.
We instead propose post-training controllable sparse \emph{video} generators, triggered by prompt keywords and emitting fixed-duration sparse clips.
This enables explicit modeling of different sparsity levels through text prompts; details are in \appref{sparse-gen}.

\subsection{Semantic representation, not temporal}
\label{sec:ats-semantic}

ATS's tree is a strictly temporal hierarchy: every node spans a contiguous time interval, and children partition the parent's interval.
This is wrong once content includes scene cuts, character recurrences, or shared environments across shots---two non-adjacent leaves may share identity, environment, or object presence while two adjacent leaves share nothing.
The right structure is a directed acyclic graph (DAG) whose nodes remain temporal spans but whose edges encode shared semantic structure rather than shared temporal nesting.

A \emph{subgraph} is a set of nodes sharing a semantic constraint: an \emph{appearance subgraph} containing every node in which entity $X$ appears (regardless of when), a \emph{shot subgraph} for every node within a continuous shot, a \emph{scene subgraph} for every shot within a setting, and so on.
Generation against such a DAG requires anchors that carry semantic identity, not just temporal-slot identity, so a leaf can be conditioned on its bracketing anchors \emph{and} on the constraints active over its span---without those constraints needing to thread through a strictly temporal parent chain.
The sparse-generation specialist of \secref{ats-t2v} is the natural place to enforce these constraints, via structured prompts in the spirit of differential text-conditioning (e.g., ``[ENVIRONMENT CHANGE: coffee shop $\to$ street],'' ``[CLOTHING CHANGE: shirt $\to$ dress]'').

This same machinery resolves limitation~3: an appearance subgraph supplies the cross-leaf identity track whose absence drives the inconsistency observed in \secref{lim-observed}, with identity constrained at the subgraph level rather than rediscovered locally at each leaf.
ATS's mechanics carry over---leaves remain bracketed by anchor pairs and run in parallel, and the latency bound scales as the depth of the DAG.
Only the topology above the leaves changes.

\subsection{Solving the continuity problem with references}
\label{sec:ats-references}

The anchors that bracket every leaf in \Eqref{tree-factor} are, in effect, short references placed at chosen times: each pins character identity, environment, and object presence locally, and the leaf generator interpolates between bracketing pairs.
References---explicit identity, environment, or appearance images and clips fed alongside the prompt---are the most robust mechanism for enforcing the continuity dimensions of \secref{introduction}, because they sidestep the suffix-based memory of \Eqref{ar-chunk} and replace it with a fixed conditioning term that does not drift.

The broader trend points the same way: recent systems are increasingly built around reference-based conditioning rather than pure prompt-based generation, with Seedance~2.0~\cite{seedance2} the most convincing case study for the power of references in T2V.
Reference-based conditioning and ATS-style bidirectional composition are complementary, not competing---a model that accepts reference inputs serves as a stronger primitive for ATS to leverage in delivering consistency alongside coherence.

\section{Conclusion}
\label{sec:conclusion}

We presented \emph{Anchored Tree Sampling} (ATS), a training-free sampler for long-horizon offline V2V generation.
ATS reorganizes long-form synthesis into one conditioning-only root call followed by a coarse-to-fine cascade of bidirectional refinement and dense leaf calls (Algorithm~\ref{alg:tree}, \Eqref{tree-factor}), replacing the $K$ sequential rollout steps of chunked AR with $L+1$ critical-path steps ($L = O(\log T)$), with sibling calls within a level conditionally independent and parallelizable across devices.

On five $30$-minute videos with Wan~$2.1$~$+$~VACE, ATS dominates two contemporary autoregressive baselines on aggregate VBench quality (AQ, IQ) and on per-chunk drift and reset-jump metrics that isolate the jumpy failure mode of AR cache resets.
It is also faster: $5.3\times$ over the strongest AR baseline at the $2000$\,s horizon.
Pushing further, ATS on LTX-$2.3$ produces single-pass, no-drift V2V outputs that exceed $40$ minutes across all five modalities---to our knowledge the first non-avatar framework whose coherent-output duration approaches the hour scale.

The path beyond ATS's current static-camera scope has three pieces.
(1) Replace the base-model root with a varying-sparsity generator so the continuity prior no longer breaks anchor emission in dynamic and T2V regimes.
(2) Generalize the strictly temporal tree to a DAG indexed by semantic subgraphs, so identity, environment, and shot-level constraints can cut across non-adjacent leaves.
(3) Strengthen both specialists with reference inputs to unlock consistency across the full duration.
With these in place, ATS becomes a general scheduler for long-form video generation without changing its parallel-sibling topology or latency bound.

ATS may also extend to soft-real-time settings by streaming leaves to the user from a populated cache of anchors, in the spirit of LPM's idle/listen/speaking states~\cite{lpm}.
Fully interactive settings, where future conditioning is unknown, would require a hybrid of ATS and autoregressive sampling and are left to future work.

\bibliographystyle{ieeetr}
\bibliography{references}

\appendix
\clearpage

\section{Additional figures} \label{app:additional_figs}
\subsection{Full-duration reels}
\begin{figure}[h]
\centering
\includegraphics[width=\columnwidth]{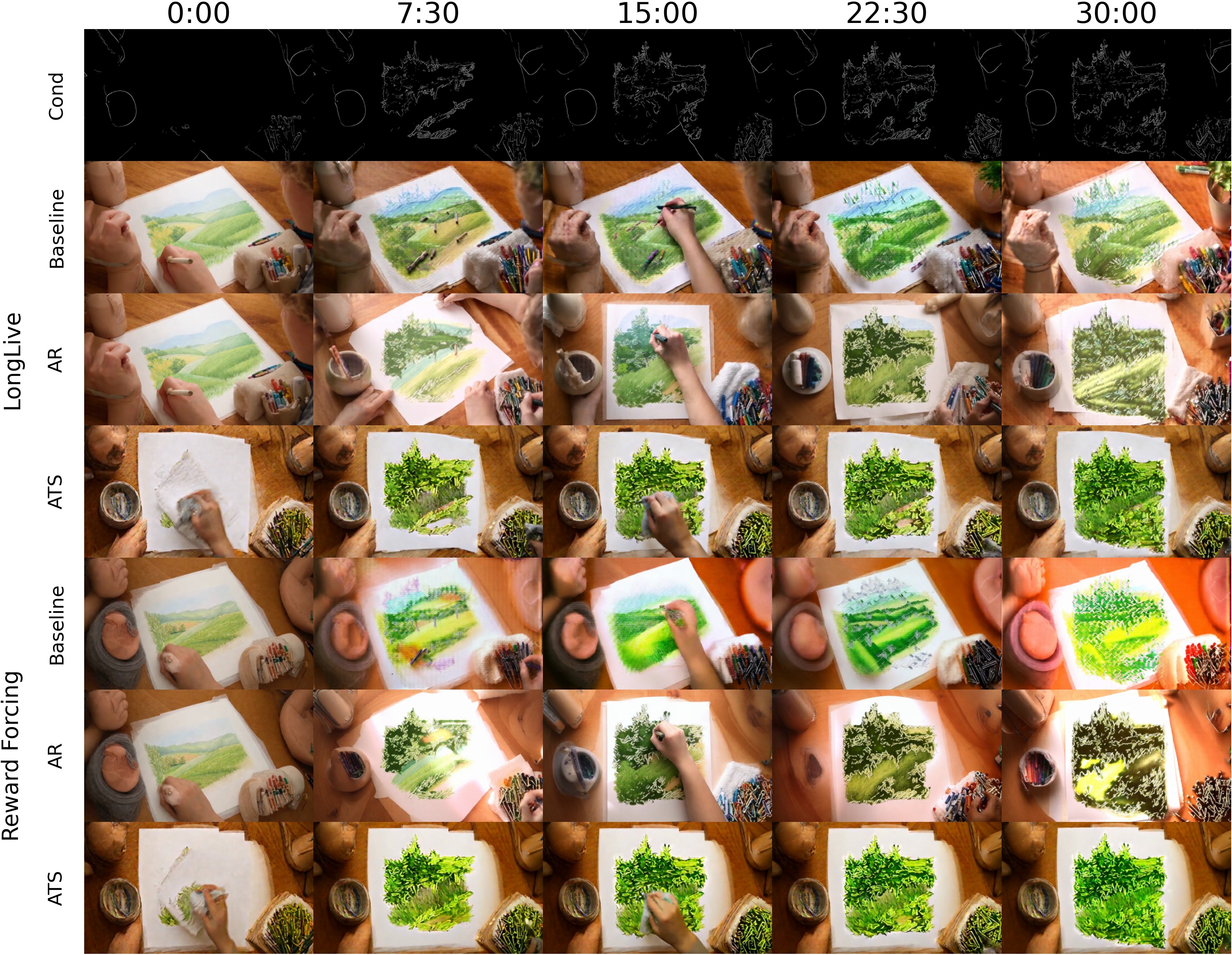}
\caption{Long-form edge-conditioned generation on Wan~$2.1$~$+$~VACE: AR vs.\ ATS on the same checkpoints.}
\label{fig:canny}
\end{figure}

\begin{figure}[h]
\centering
\includegraphics[width=\columnwidth]{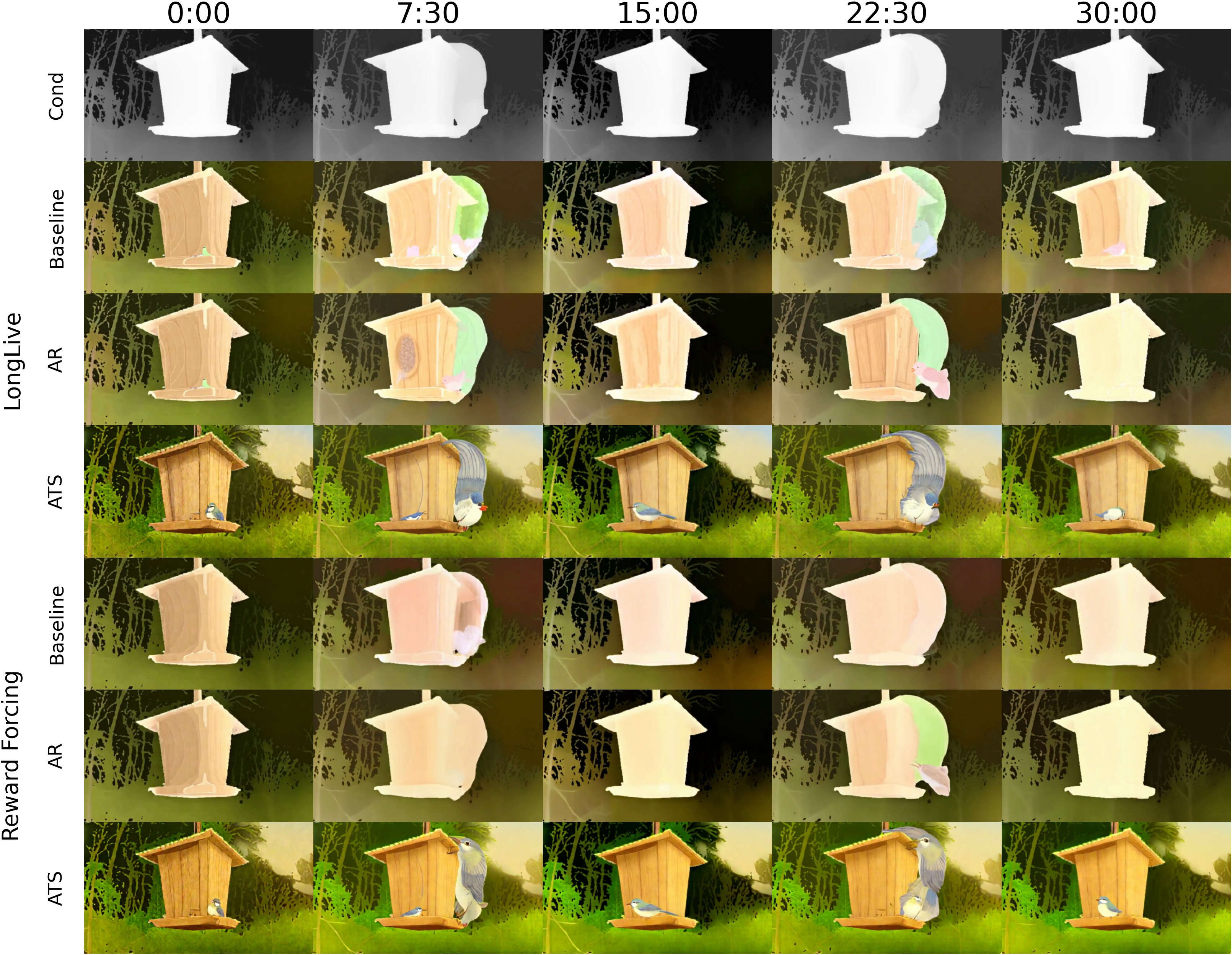}
\caption{Long-form depth-conditioned generation on Wan~$2.1$~$+$~VACE: AR vs.\ ATS on the same checkpoints.}
\label{fig:depth}
\end{figure}

\begin{figure}[h]
\centering
\includegraphics[width=\columnwidth]{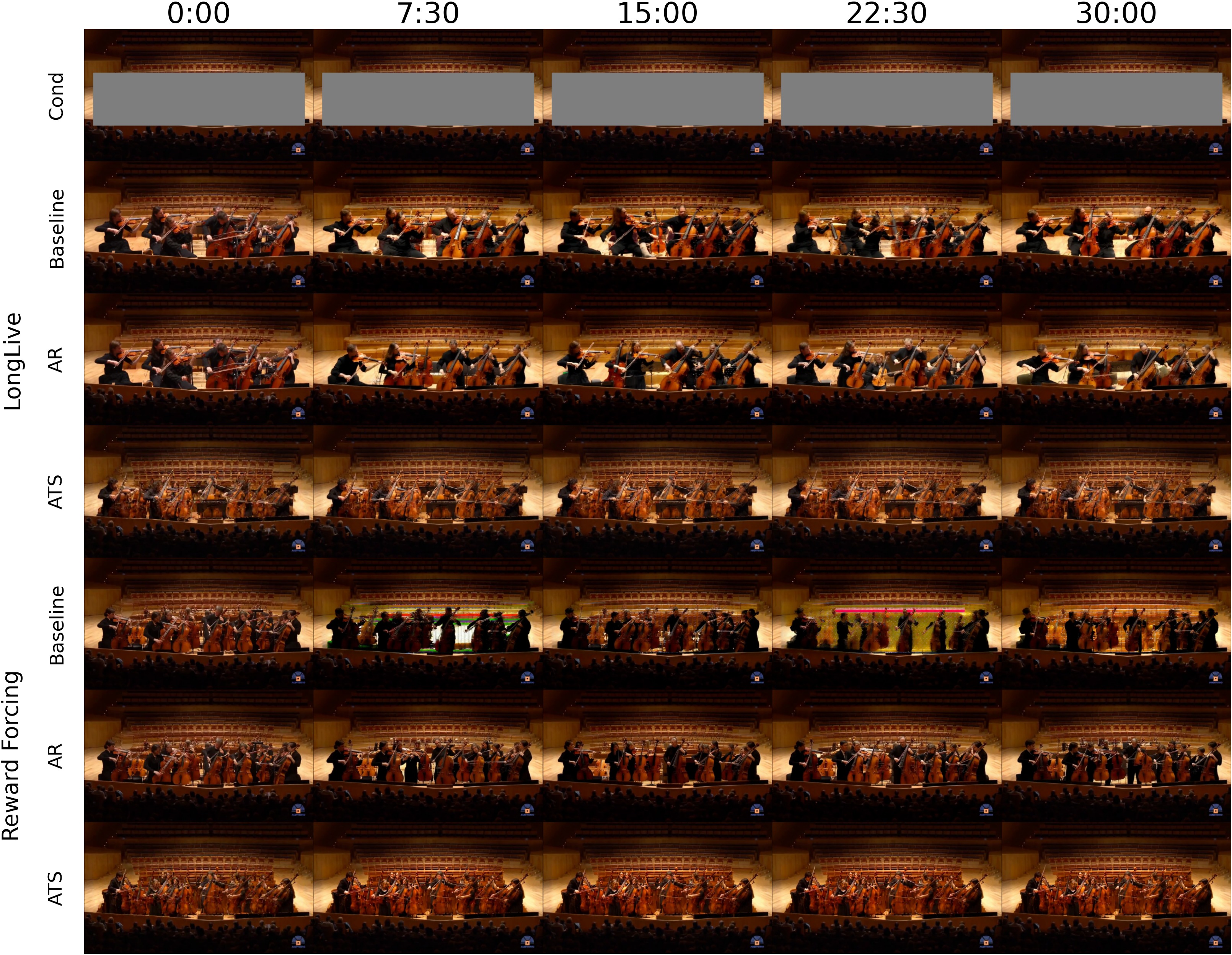}
\caption{Long-form inpainting generation on Wan~$2.1$~$+$~VACE: AR vs.\ ATS on the same checkpoints.}
\label{fig:inpaint}
\end{figure}

\begin{figure}[h]
\centering
\includegraphics[width=0.8\columnwidth]{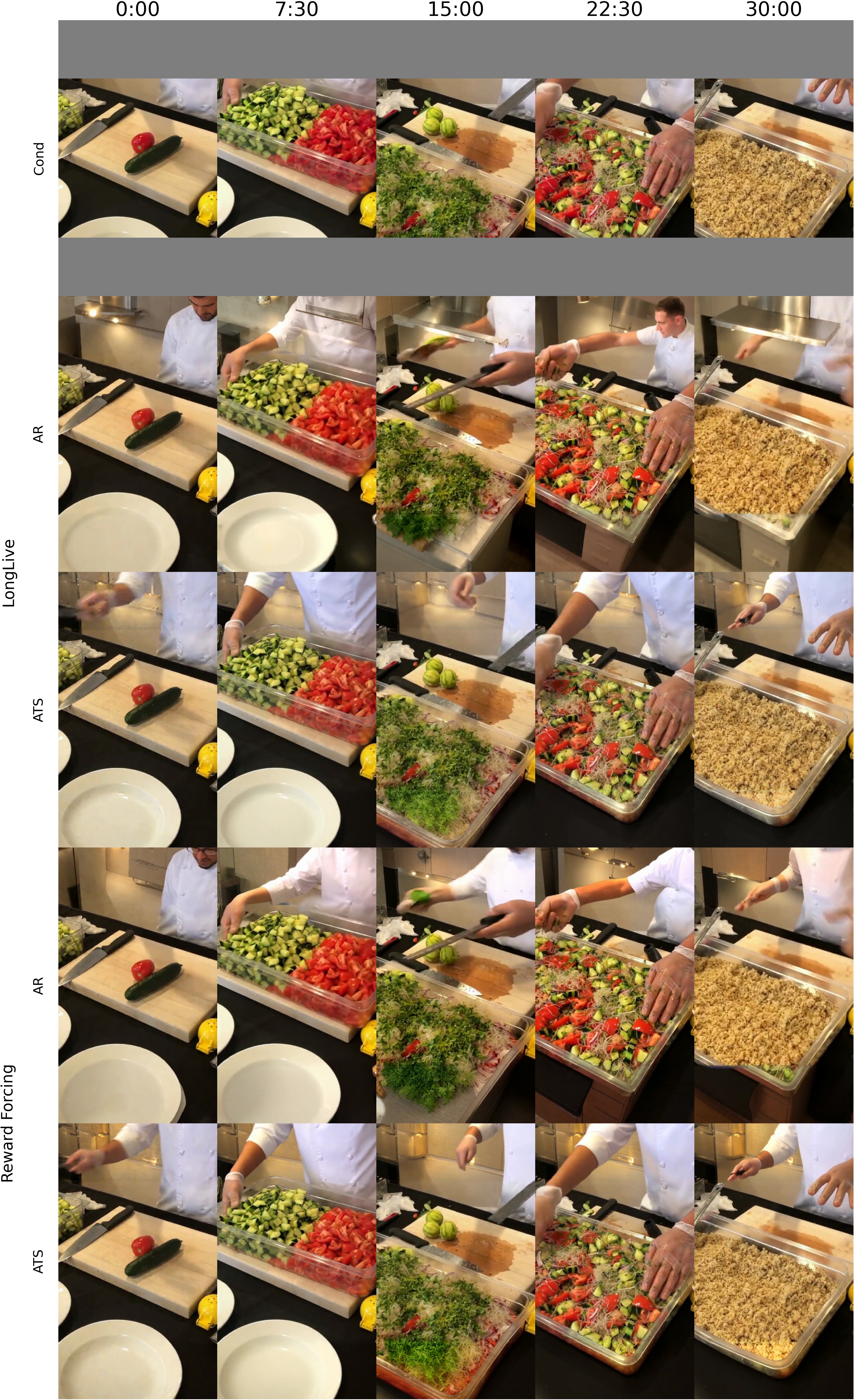}
\caption{Long-form outpainting on Wan~$2.1$~$+$~VACE: AR vs.\ ATS on the same checkpoints.}
\label{fig:outpaint}
\end{figure}

\clearpage
\subsection{Drift reels}
\begin{figure}[h]
\centering
\includegraphics[width=\columnwidth]{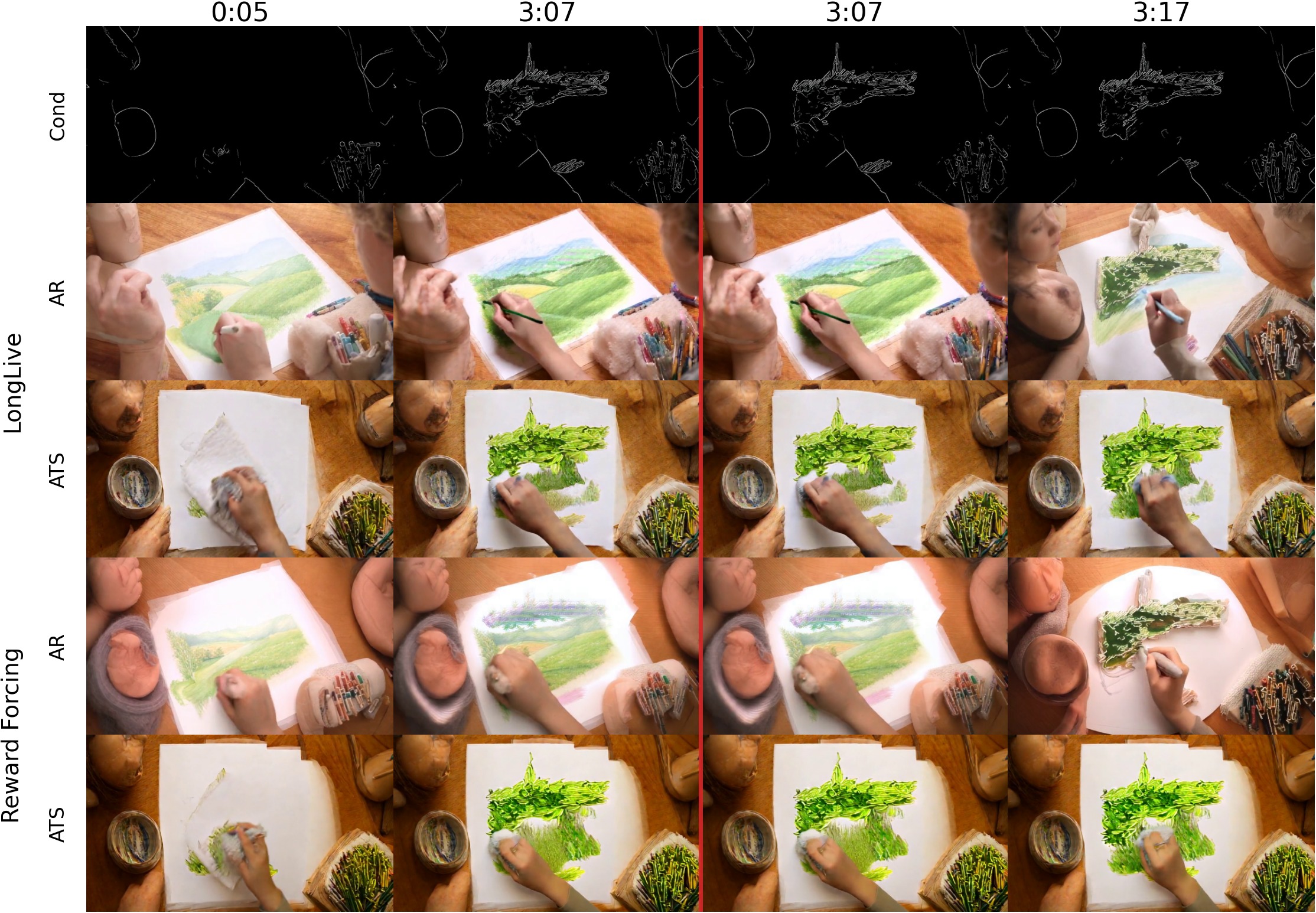}
\caption{Qualitative drift and cache-reset visualization for edge-guided Wan~$2.1$~$+$~VACE generation. The left pair shows intra-chunk drift accumulated within a single cache window, while the right pair illustrates the cross-reset jump.}
\label{fig:cache_canny}
\end{figure}

\begin{figure}[t]
\centering
\includegraphics[width=\columnwidth]{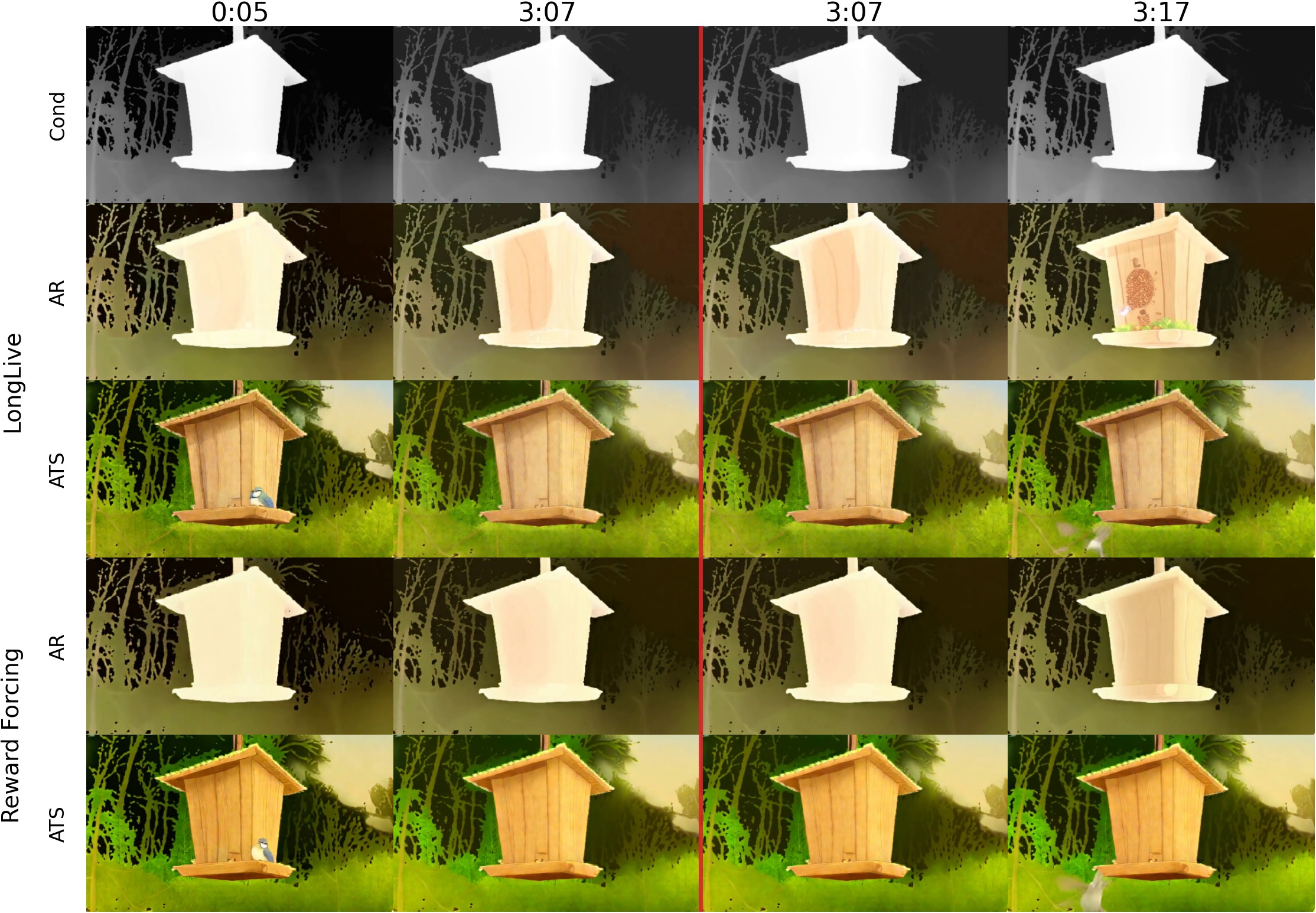}
\caption{Qualitative drift and cache-reset visualization for depth-guided Wan~$2.1$~$+$~VACE generation. The left pair shows intra-chunk drift accumulated within a single cache window, while the right pair illustrates the cross-reset jump.}
\label{fig:cache_depth}
\end{figure}

\begin{figure}[t]
\centering
\includegraphics[width=\columnwidth]{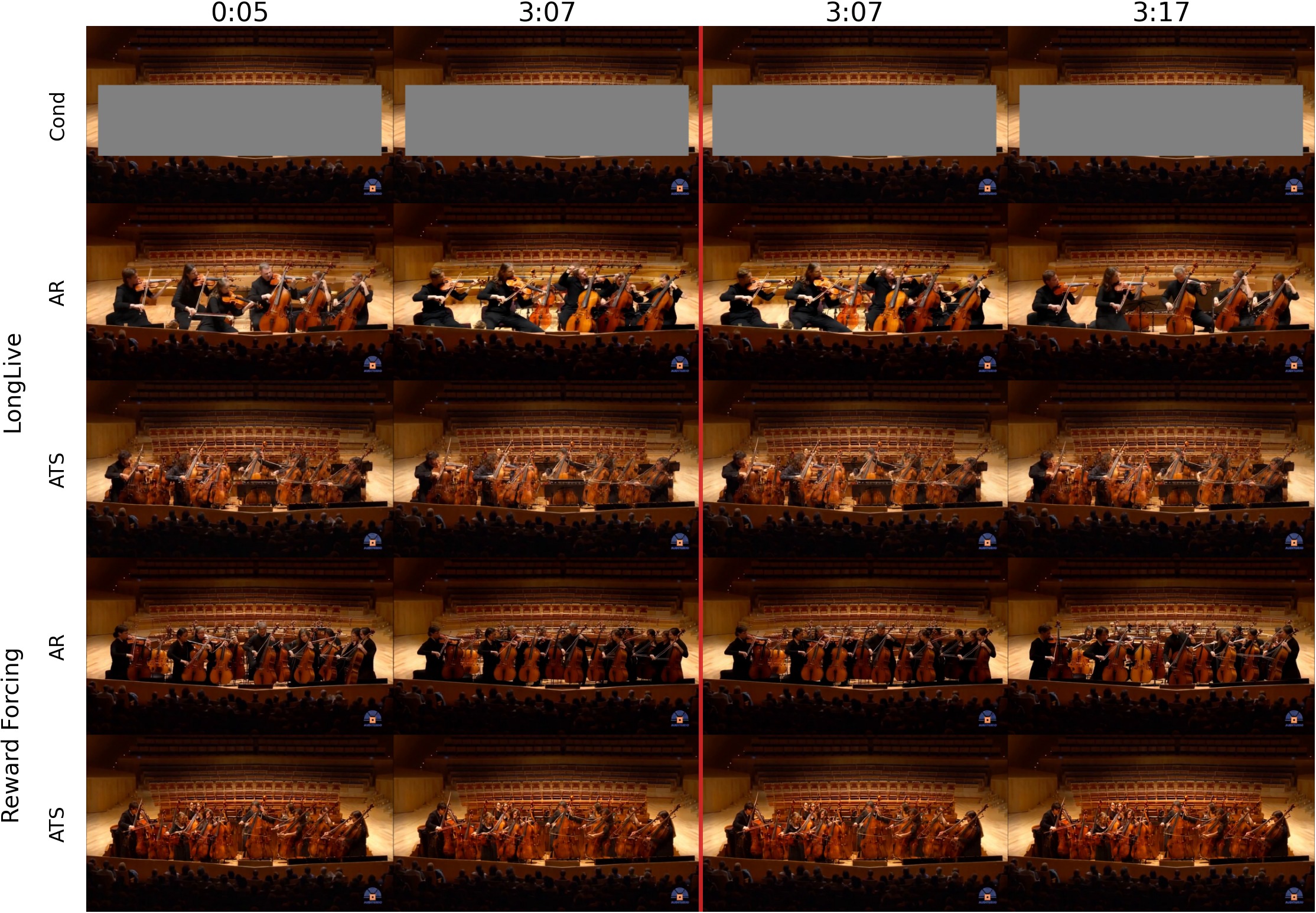}
\caption{Qualitative drift and cache-reset visualization for Wan~$2.1$~$+$~VACE inpainting. The left pair shows intra-chunk drift accumulated within a single cache window, while the right pair illustrates the cross-reset jump.}
\label{fig:cache_inpaint}
\end{figure}

\begin{figure}[t]
\centering
\includegraphics[width=\columnwidth]{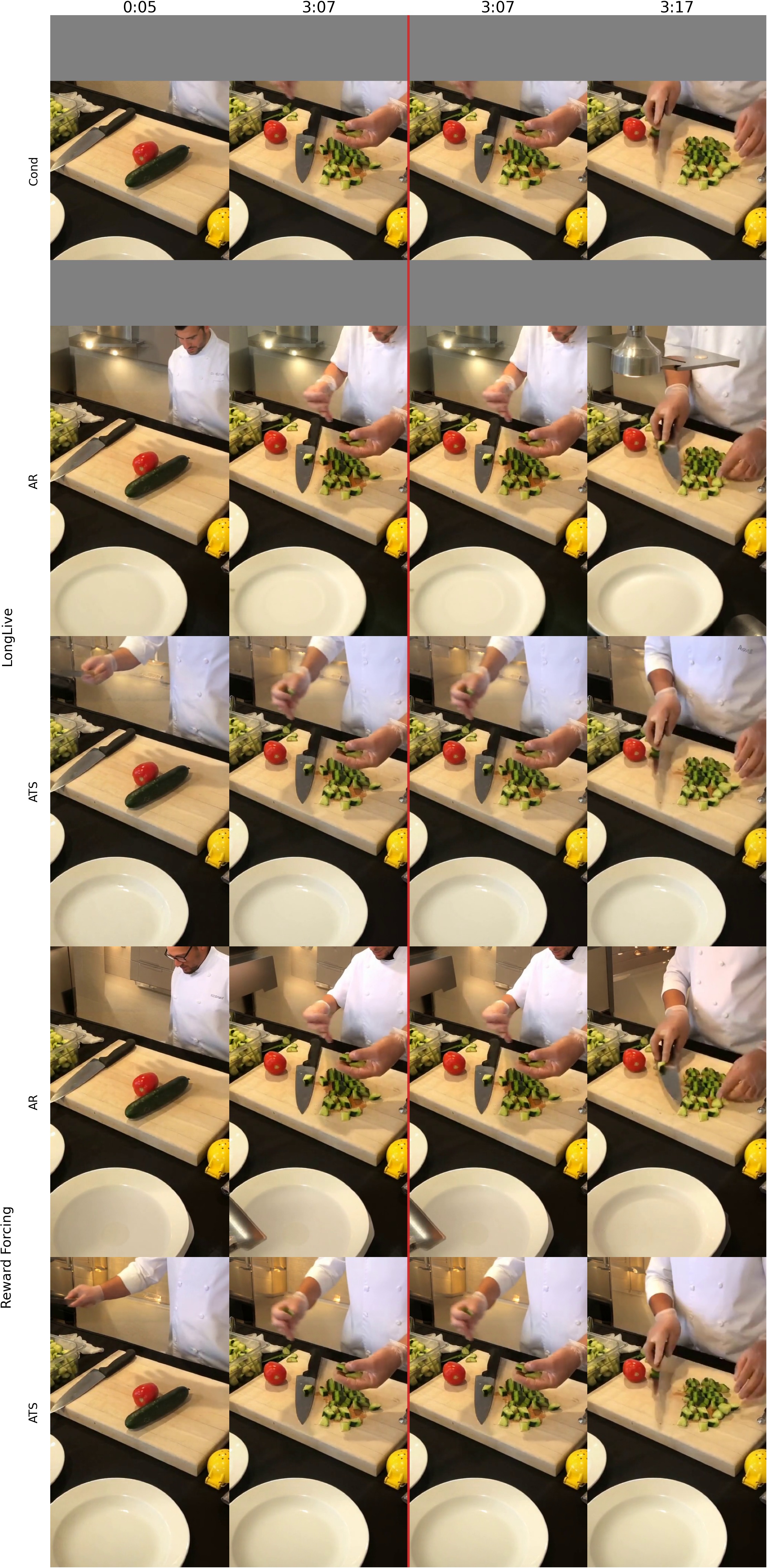}
\caption{Qualitative drift and cache-reset visualization for Wan~$2.1$~$+$~VACE outpainting. The left pair shows intra-chunk drift accumulated within a single cache window, while the right pair illustrates the cross-reset jump.}
\label{fig:cache_outpaint}
\end{figure}

\clearpage
\section{Training a sparse video generator}
\label{app:sparse-gen}

This appendix sketches the training strategy for the sparse-generation specialist that powers the ATS root call beyond the static-camera regime (\secref{ats-t2v}).
The intent is to fix a concrete design we believe is sufficient, not to specify every implementation detail.
The latter is left for future work.

\paragraph{Why post-train rather than train from scratch?}
Pretrained video generators have already learned what they need about appearance, motion, and short-range temporal structure.
What they lack is the ability to emit \emph{sparse} content on demand.
We therefore propose a LoRA-based post-training framework on top of a strong base model such as LTX-$2$~\cite{ltx2}.
The goal is to teach the model to recognize a small vocabulary of \emph{trigger tokens} in the prompt and to emit appropriately sparse output when those tokens appear.
This is coupled with fixed lengths for each sparse clip the model produces to ensure reliability (e.g., 24 frames per clip).

\paragraph{Trigger tokens for sparsity.}
The trigger tokens form a hierarchy of \emph{how much carries over} between adjacent chunks of the sparse output.
Four tokens cover the full range, plus one modifier:

\begin{itemize}
\item \texttt{[TIMESKIP]}: same shot, same setting, same subject; only time advances.
Used for sparse temporal sampling within a single continuous take.
This is the \emph{intra-shot} level of the sparsity ladder in \secref{ats-t2v}.

\item \texttt{[SHOT CHANGE]}: different shot, same scene.
Camera angle, framing, and possibly subject focus shift; the setting persists.
This is the \emph{intra-scene} level.

\item \texttt{[SCENE CHANGE]}: the same subject reappears in a different scene.
Identity persists; location, time-of-day, and lighting change.
The model is asked to render the same character recognizably in a new context.

\item \texttt{[NEW SUBJECT]}: a fully new subject in a fully new scene.
Nothing carries over visually.
This is the \emph{intra-full-video} level, and also the right token when the sparse generator opens a new narrative thread.

\item \texttt{[STATIC CAMERA]}: a modifier (not a transition) indicating that the camera does not move during the segment.
Provides explicit control over the regime where ATS already works today.
\end{itemize}

A multi-segment prompt is then a sequence of captions separated by transition tokens, of the form
\texttt{[1] $\langle$caption\textsubscript{1}$\rangle$ [TIMESKIP] [2] $\langle$caption\textsubscript{2}$\rangle$ [SHOT CHANGE] [3] $\langle$caption\textsubscript{3}$\rangle\,\cdots$}.
At inference time, the token sequence \emph{is} the sparsity specification.
A prompt with mostly \texttt{[TIMESKIP]} tokens asks for a long single take seen through a handful of representative frames, a prompt with mostly \texttt{[SCENE CHANGE]} tokens asks for representative frames spanning multiple scenes of the same character, and a mix yields a multi-scene, multi-shot, multi-subject sparse layout.

\paragraph{Modeling sparsity through text rather than architecture.}
The key design decision is that \emph{sparsity is selected via prompt, not via a separate head, channel, or routing mechanism}.
The same model, on the same forward pass, can produce dense intra-shot content (only \texttt{[TIMESKIP]} transitions), semi-sparse intra-scene content (mixed \texttt{[TIMESKIP]} and \texttt{[SHOT CHANGE]}), or maximally sparse intra-full-video content (\texttt{[SCENE CHANGE]} and \texttt{[NEW SUBJECT]}), just by varying the token sequence.
This makes the sparse generator drop-in compatible with the ATS tree of \secref{tree}, where the root call uses a mostly-\texttt{[SCENE CHANGE]}\,/\,\texttt{[NEW SUBJECT]} prompt to produce $\mc{A}^{(0)}$ over the entire horizon, refinement levels use less-sparse prompts (mostly \texttt{[SHOT CHANGE]}), and leaves are still rendered by the unmodified dense base model with no trigger tokens at all.
The sparsity ladder in \secref{ats-t2v} is then literally a ladder of token mixes.

This is also the natural place to host the differential text-conditioning constructs of \secref{ats-semantic}.
Compound prompts of the form \texttt{[SCENE CHANGE: coffee shop $\to$ street]} or \texttt{[SHOT CHANGE: trailing person $\to$ over-the-shoulder]} attach the specific \emph{delta} to the transition token, telling the model what changes while everything else (identity, wardrobe, time-of-day) carries over by default.

\paragraph{Training data: matching each rung of the ladder to a corpus.}
Each transition token corresponds to a different pairing structure in the training data, and the natural data source differs by rung:

\begin{itemize}
\item \texttt{[TIMESKIP]} pairs are easy to construct from any continuous video clip by sparsely subsampling frames within a single shot.
Large-scale clip datasets such as OpenVid-1M~\cite{openvid} suffice for this rung.

\item \texttt{[SHOT CHANGE]} pairs require within-scene multi-shot data: the same setting, a different camera, sometimes a different subject focus.
Shot-segmented documentary and film footage covers this.

\item \texttt{[SCENE CHANGE]} pairs are the hardest to source: the same subject must reappear across genuinely different scenes (location, time-of-day, lighting), with identity preserved.
This is the rung where long-form narrative data becomes essential.

\item \texttt{[NEW SUBJECT]} pairs are trivially abundant: any two unrelated clips form a valid pair.
\end{itemize}

The natural choice for the upper rungs of the ladder is MovieBench~\cite{moviebench}, a hierarchical movie-level dataset for long-video generation that explicitly indexes recurring characters across scenes within feature-length narratives.
MovieBench supplies precisely the same-subject / different-scene structure that \texttt{[SCENE CHANGE]} needs to learn, and the multi-scene structure of feature-length films naturally exposes the full sparsity ladder \emph{within a single source}.
Temporally related content (recurring characters, persistent environments, intra-scene shot variety) and temporally unrelated content (cuts between unrelated scenes) live side by side in the same corpus.
Pairing OpenVid for the dense intra-shot end with MovieBench for the very-sparse intra-full-video end covers the entire ladder with off-the-shelf data.

\paragraph{Composition with ATS.}
The sparse generator is a drop-in replacement for the conditioning-only root call of \secref{tree-inference}.
Concretely, instead of feeding $\vec{c}_{1:T}$ to an unmodified base model and asking for sparse output (which fails outside the static-camera regime), the planner emits a token-structured prompt to the sparse generator, which produces $\mc{A}^{(0)}$ at the chosen anchor times.
From there, ATS proceeds exactly as in \secref{tree-inference}.
The latency bound \Eqref{latency}, the parallel-sibling topology, and the anchor-bounded drift behaviour of ATS all carry over unchanged.
The only thing the sparse generator changes is whether $\mc{A}^{(0)}$ is recoverable in non-static regimes, which is the entire purpose of the addition.

\clearpage
\section{LTX-2.3 full-duration reels}
\label{app:ltx-full}

We evaluated the same five videos considered in \secref{experiments} for the same tasks with LTX-$2.3$ \cite{ltx2}. This appendix collects the LTX-$2.3$ per-modality reels at their full $\geq 40$-minute duration, complementing the hero figure (\Figref{hero}).
Where the body figures in \secref{experiments} are the $30$-minute Wan~$2.1$~$+$~VACE head-to-head comparisons against the autoregressive baselines, the figures below show the longer LTX-$2.3$ reels we use to demonstrate that ATS's no-quality-drift property survives well past the operating range of any current AR sampler.
The full videos can be viewed at \url{https://ats.github.io/}.

\begin{figure}[h]
\centering
\includegraphics[width=\columnwidth]{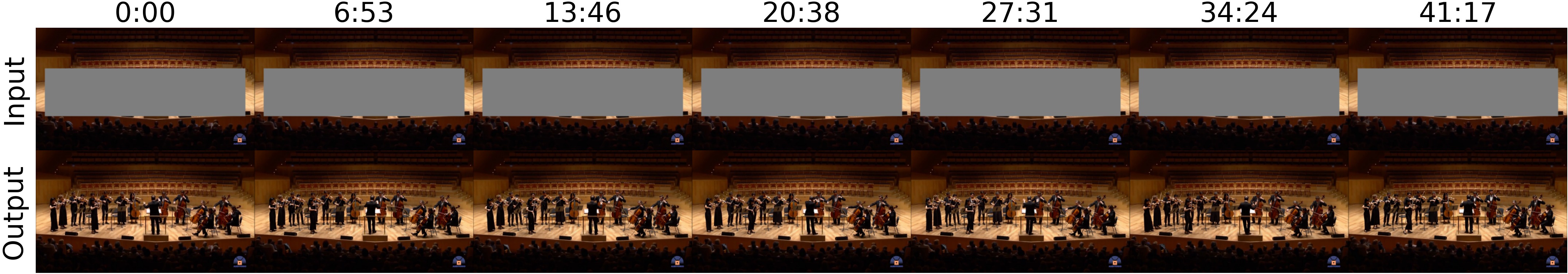}
\caption{Long-form video inpainting with ATS on LTX-$2.3$, full duration.}
\label{fig:ltx-inpaint}
\end{figure}

\begin{figure}[h]
\centering
\includegraphics[width=\columnwidth]{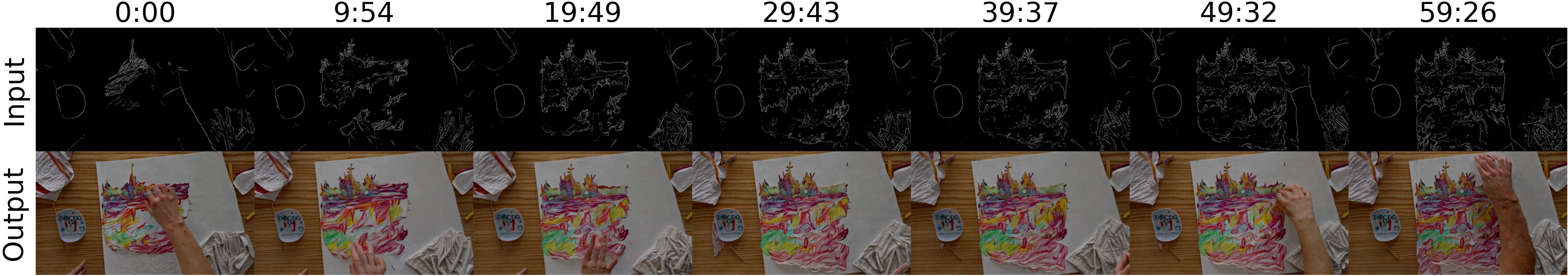}
\caption{Long-form edge-conditioned (Canny) generation with ATS on LTX-$2.3$, full duration.}
\label{fig:ltx-canny}
\end{figure}

\begin{figure}[h]
\centering
\includegraphics[width=\columnwidth]{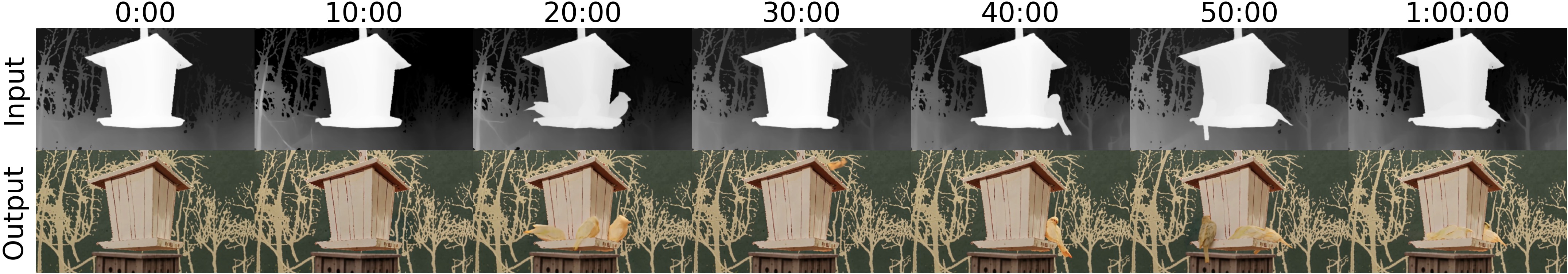}
\caption{Long-form depth-conditioned generation with ATS on LTX-$2.3$, full duration.}
\label{fig:ltx-depth}
\end{figure}

\begin{figure}[h]
\centering
\includegraphics[width=\columnwidth]{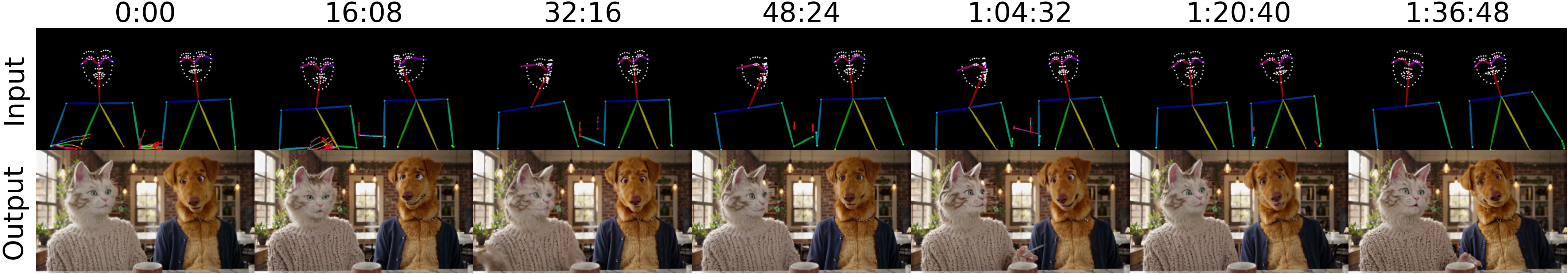}
\caption{Long-form pose-conditioned generation with ATS on LTX-$2.3$, full duration.}
\label{fig:ltx-pose}
\end{figure}

\begin{figure}[h]
\centering
\includegraphics[width=\columnwidth]{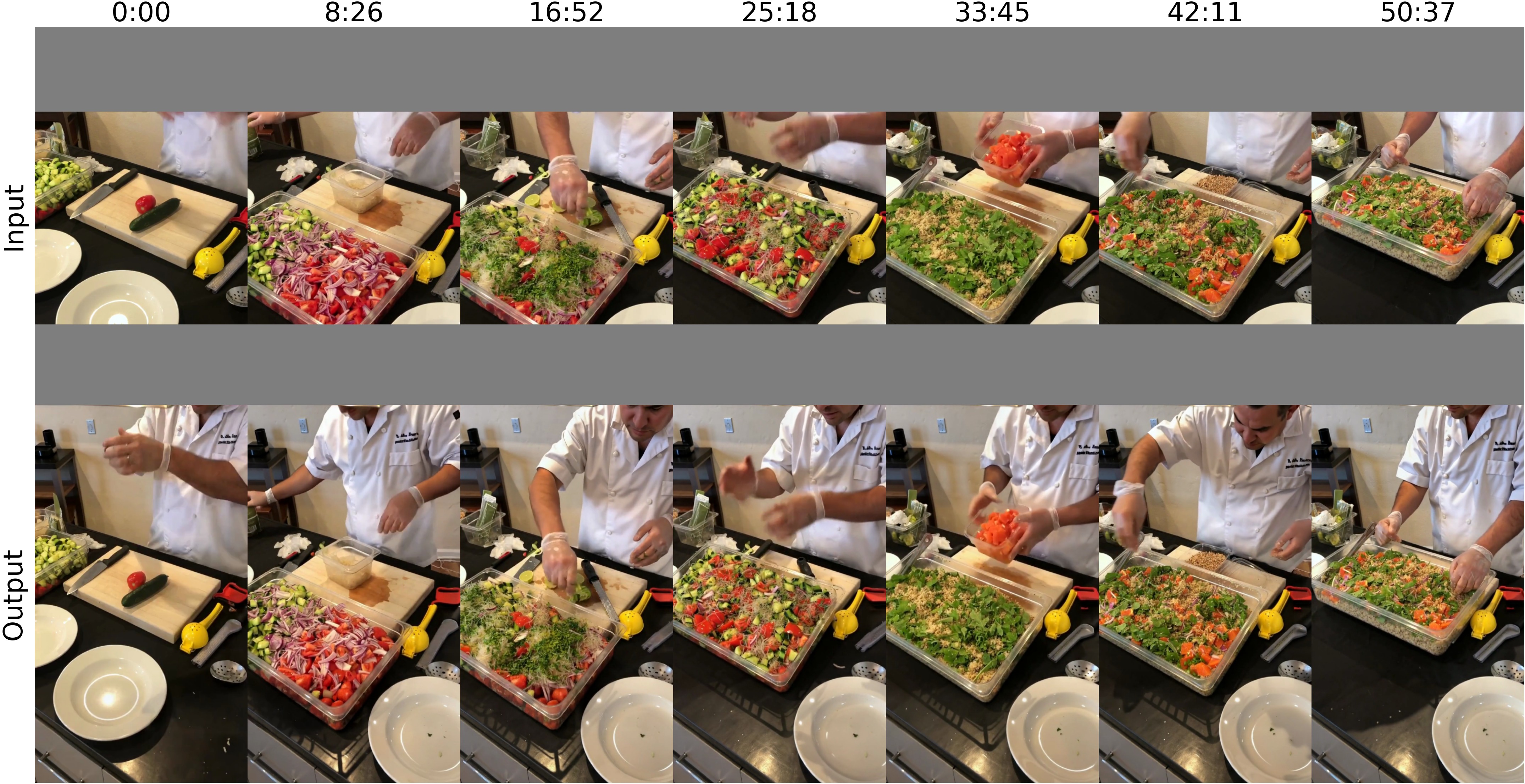}
\caption{Long-form video outpainting with ATS on LTX-$2.3$, full duration.}
\label{fig:ltx-outpaint}
\end{figure}

\clearpage
\section{Source videos}
\label{app:eval-set}

The five source videos used for the qualitative reels (\secref{experiments}, \appref{ltx-full}) and the head-to-head VBench comparison (\tabref{vbench}) are listed below.
Each clip is trimmed to a $30$-minute window for the Wan~$2.1$~$+$~VACE evaluation in the body, and to a $\geq 40$-minute window for the LTX-$2.3$ reels in \appref{ltx-full}.
The table provides YouTube URLs sufficient to reproduce the benchmark from public sources.
We do not redistribute the source media.

\begin{table}[h]
\centering
\caption{Source videos. One clip per VACE conditioning modality.}
\label{tab:source-videos}
\small
\setlength{\tabcolsep}{6pt}
\begin{tabular}{ll}
\toprule
Modality & URL \\
\midrule
Inpainting   & \url{https://www.youtube.com/watch?v=4rgSzQwe5DQ} \\
Outpainting  & \url{https://www.youtube.com/watch?v=wfYHoSzxk_o} \\
Edge (Canny) & \url{https://www.youtube.com/watch?v=e84sUd_QM6E} \\
Depth        & \url{https://www.youtube.com/watch?v=61P9wlLyME8} \\
Pose         & \url{https://www.youtube.com/watch?v=LeYIndII13w} \\
\bottomrule
\end{tabular}
\end{table}

\end{document}